%% file: LMGAE-UAI.tex
\documentclass[letterpaper]{article}
\usepackage{uai2020}
\usepackage[margin=1in]{geometry}
\pdfoutput=1
\usepackage{times}

\usepackage{graphicx}
\usepackage{amsmath}
\usepackage{amssymb}
\usepackage{amsthm}
\usepackage{xcolor}
\usepackage{multirow}
\usepackage{algpseudocode,algorithm,algorithmicx}
\usepackage{natbib}
\newtheorem{theorem}{Theorem}
\input{math_commands.tex}

\newcommand*\Let[2]{\State #1 $\gets$ #2}
\newcommand{\printfnsymbol}[1]{%
  \textsuperscript{\@fnsymbol{#1}}%
}
\algrenewcommand\algorithmicrequire{\textbf{Precondition:}}
\algrenewcommand\algorithmicensure{\textbf{Postcondition:}}

\newcommand{\real}[1]{\displaystyle \R^{#1}}
\newcommand{\encoder}{$E_{\kappa}$ }
\newcommand{\decoder}{$D_{\psi}$ }
\newcommand{\discriminator}{$H_{\zeta}$ }

\newcommand{\centered}[1]{\begin{tabular}{l} #1 \end{tabular}}

\definecolor{violet}{rgb}{0,1,0}
\definecolor{blue}{rgb}{0,0,0}
\title{MaskAAE: Latent space optimization for Adversarial Auto-Encoders}
\author{Arnab Kumar Mondal\thanks{.  these authors contributed equally} \\
IIT Delhi\\
{\small anz188380@cse.iitd.ac.in}
\And
Sankalan Pal Chowdhury\footnotemark[1]\\
IIT Delhi\\
{\small cs1160701@iitd.ac.in}
\And
Aravind Jayendran \footnotemark[1]\\
Flipkart Internet Pvt. Ltd.\thanks{.  work done when at IIT Delhi}\\
{\small aravind.j@flipkart.com}
\AND
Parag Singla\\
IIT Delhi\\
{\small parags@cse.iitd.ac.in}
\And
Himanshu Asnani\\
TIFR\\
{\small himanshu.asnani@tifr.res.in}
\And
Prathosh AP\\
IIT Delhi\\
{\small  prathoshap@ee.iitd.ac.in}
}

\begin{document}

\maketitle

%%%%%%%%% ABSTRACT
\input{abstract}
%%%%%%%%% BODY TEXT
\input{intro}
\input{related}
\input{theory}
\input{model}

\input{expt}

\input{discussion}

{\small
\bibliographystyle{IEEEtranSN}
\bibliography{egbib}
}
\newpage
\input{supp_theory}
\input{supp_model}
\input{supp_expt}

\end{document}

%% file: math_commands.tex
\usepackage{amsmath,amsfonts,bm}

\def\eqref#1{equation~\ref{#1}}
\def\1{\bm{1}}

\def\rz{{\textnormal{z}}}

\def\vu{{\bm{u}}}

\def\vx{{\bm{x}}}

\def\vz{{\bm{z}}}

\DeclareMathAlphabet{\mathsfit}{\encodingdefault}{\sfdefault}{m}{sl}
\SetMathAlphabet{\mathsfit}{bold}{\encodingdefault}{\sfdefault}{bx}{n}

\newcommand{\R}{\mathbb{R}}

%% file: abstract.tex
\begin{abstract}

The field of neural generative models is dominated by the highly successful Generative Adversarial Networks (GANs) despite their challenges, such as training instability and mode collapse. Auto-Encoders (AE) with regularized latent space provide an alternative framework for generative models, albeit their performance levels have not reached that of GANs. In this work, we hypothesise that the dimensionality of the AE model's latent space has a critical effect on the quality of generated data. Under the assumption that nature generates data by sampling from a ``true" generative latent space followed by a deterministic  function,  we show that the optimal performance is obtained when the dimensionality of the latent space of the AE-model matches with that of the ``true" generative latent space. Further, we propose an algorithm called the Mask Adversarial Auto-Encoder (MaskAAE), in which the dimensionality of the latent space of an adversarial auto encoder is brought closer to that of the ``true" generative latent space, via a procedure to mask the spurious latent dimensions. We demonstrate through experiments on synthetic and several real-world datasets that the proposed formulation yields betterment in the generation quality.
\end{abstract}

%% file: intro.tex
\section{INTRODUCTION}\label{sec:intro}

The objective of a probabilistic generative model is to learn to sample new points from a distribution given a finite set of data points drawn from it. Deep generative models, especially the Generative Adversarial Networks (GANs) (\cite{NIPS2014_5423}) have shown remarkable success in this task by generating high quality data (\cite{brock2018large}). GANs implicitly learn to sample from the data distribution by transforming a sample from a simplistic distribution (such as Gaussian) to the sample from the data distribution by optimising a min-max objective through an adversarial game between a pair of function approximators called the generator and the discriminator. Although GANs generate high-quality data, they are known to suffer from problems like instability of training (\cite{arora2017generalization,salimans2016improved}), degenerative supports for the generated data (mode collapse) (\cite{arjovsky2017towards,srivastava2017veegan}) and sensitivity to hyper-parameters (\cite{brock2018large}). 

Auto-Encoder (AE) based generative models (\cite{zhao2017infovae,kingma2013autoencoding,44904,46498}) provide an alternative to GAN based models. The fundamental idea is to learn a lower dimensional latent representation of data through a deterministic or stochastic encoder and learn to generate (decode) the data through a decoder. Typically, both the encoder and decoder are realised through learnable family of function approximators or deep neural networks. To facilitate the generation process, the distribution over the latent space is forced to follow a known distribution so that sampling from it is feasible. Despite resulting in higher data-likelihood and stable training, the quality of generated data of the AE-based models is known to be far away from state-of-the-art GAN models (\cite{dai2019diagnosing,grover2017flow,theis2015note}).\par While there have been several angles of looking at the shortcomings of the AE-based models (\cite{dai2019diagnosing,hoshen2019non, kingma2016improved, tomczak2017vae,klushyn2019learning,bauer2018resampled,van2017neural}), an important question seems to have remained unaddressed: How does the dimensionality of the latent space (bottle-neck layer) affect the generation quality in AE-based models?

It is a well-known fact that most of the naturally occurring data effectively lies in a manifold with dimension much lesser than its original dimensionality (\cite{cayton2005algorithms,law2006incremental,narayanan2010sample}). Intuitively, this suggests that with functions that Deep Neural Networks learn, there exists an optimal number of latent dimensions, since ``lesser" or ``extra" number of latent dimensions  may result in loss of information and noisy generation, respectively. This observation is also corroborated by empirical evidence provided in Fig. \ref{fig:ucurve} where a state-of-the-art AE-based generative model (Wasserstein Auto-Encoder \cite{zhang2019wassersteinwasserstein}) is constructed on two synthetic (detailed in Section 5) and MNIST datasets, with varying latent dimensionality (everything else kept the same). It is seen that the generation quality metric (FID) follows a U-shaped curve. Thus, to obtain optimal generation quality, a brute-force search over a large range of values of latent dimensionality may be required, which is practically infeasible.  Motivated by the aforementioned observations, in this work, we explore the role of latent dimensionality in AE-based generative models, with the following contributions:

\begin{figure}
    \begin{center}
        \includegraphics[keepaspectratio, width=\columnwidth]{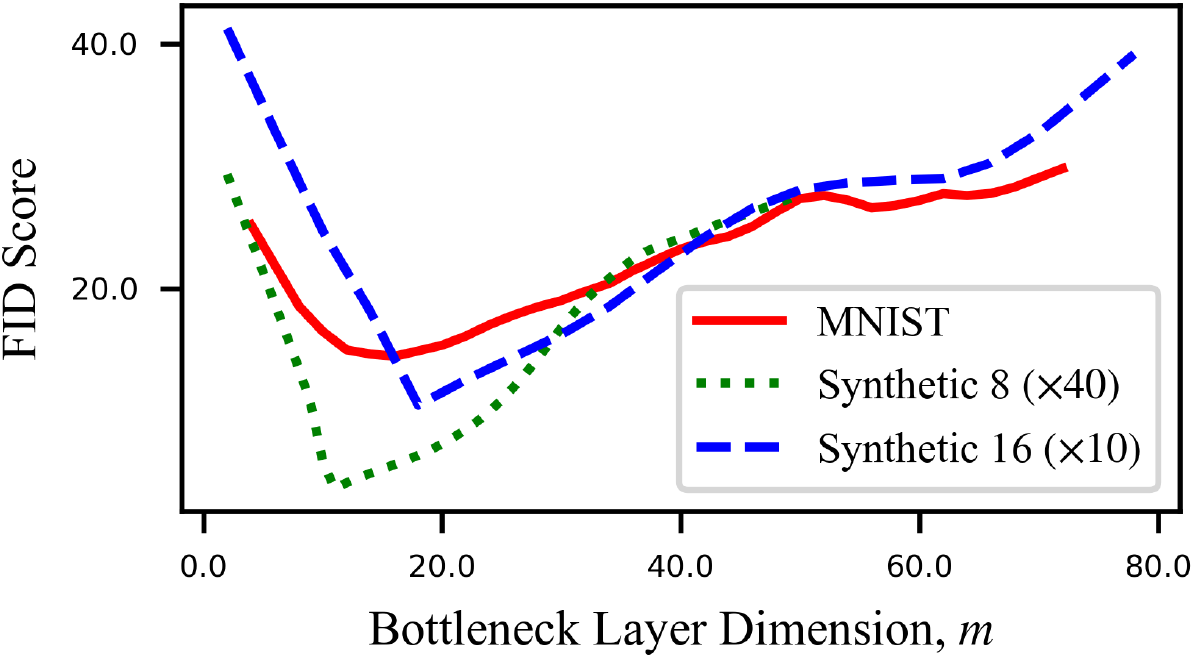}
    \end{center}
    \vspace{-4mm}
    \caption{FID score for a Wasserstein Auto-Encoder with varying latent dimensionality $m$  for 2 synthetic datasets of `true' latent dimensions, $n=8$  and $n=16$ and MNIST. It is seen that the generation quality gets worse on both the sides of a certain latent dimensionality. FID scores have been scaled appropriately to bring them in the same range}
    \vspace{-2mm}
    \label{fig:ucurve}
\end{figure}

\begin{enumerate}

\item We model the data generation as a two-stage process comprising of sampling from a ``true'' latent space followed by a deterministic function.% and show that under this model, optimal generation quality is achieved by an AE-model when the latent dimensionality of the AE-model is equal ``true'' latent dimensionality that is assumed to generate data. 
\item We provide theoretical understanding on the role of the dimensionality of the latent space on the generation quality, by formalizing the requirements for a faithful generation in of AE-based generative models with deterministic encoder and decoder networks.  
\item Owing to the obliviousness of the dimensionality of the ``true'' latent space in real-life data, we propose a method to algorithmically ``mask'' the spurious dimensions in AE-based models (and thus call our model the MaskAAE).  
\item We demonstrate the efficacy of the proposed model on synthetic as well as large-scale image datasets by achieving better generation quality metrics compared to the state-of-the-art AE-based models. 
    
\end{enumerate}

%% file: related.tex
\section{RELATED WORK}\label{sec:related}

Let $\vx$ denote data points lying in the space $\mathcal{X}$ conforming to an underlying distribution $\Upsilon(\vx)$, from which a generative model desires to sample. An Auto-Encoder based model constructs a lower-dimensional latent space $\mathcal{Z}$ to which the data is projected through an (probabilistic or deterministic) Encoder function, $E_\kappa$. 
An inverse projection map is learned from $\mathcal{Z}$ to $\mathcal{X}$ through a Decoder function $D_\psi$, which can be subsequently used as a sampler for $\Upsilon(\vx)$. For this to happen, it is necessary that the distribution of points over the latent space $\mathcal{Z}$ is regularized (to some known distribution $\Pi(\vz)$) to facilitate explicit sampling from $\Pi(\vz)$, so that decoder can generate data taking samples from $\Pi(\vz)$ as input. Most of the AE-based models %direction proposed in  Variational Auto-Encoders (VAE) \cite{kingma2013autoencoding} where the Encoder outputs the parameters (or samples) of a conditional latent posterior $q(\vz|\vx)$ which is forced to be follow a predefined prior (e.g., Isotropic Normal distribution), through minimization of a divergence measure and the Decoder generates data by taking the samples of latent distribution as input.  This approach equivalently maximizes a 
 maximize the data likelihood (or a lower bound on it), which is shown (\cite{kingma2013autoencoding,hoffman2016elbo}) to consist of the sum of two critical terms - (i) the likelihood of the Decoder generated data %$\mathbb{E}_{\Psi(\vz|\vx)}\Big[\log\big(\Phi(\vx|\vz\big)\Big] $
and, (ii) a divergence measure between the assumed latent distribution, $\Pi(\vz)$, and the distribution imposed on the latent space by the Encoder, $\Psi(\vz)=\int {\Psi(\vz|\vx) \Upsilon(\vx) d\vx}$,   (\cite{hoffman2016elbo,44904}). This underlying commonality, suggests that the success of an AE-based generative model depends upon  simultaneously optimising the aforementioned terms.
The first criterion is fairly easily ensured in all AE models by minimizing a surrogate function such as the reconstruction error between the samples of the true data and output of the decoder, which can be made arbitrarily small (\cite{burgess2018understanding,dai2019diagnosing,alain2014regularized}) by increasing the network capacity. It is well recognized that the quality of the generated data relies heavily on achieving the second criteria of bringing the Encoder imposed latent distribution $\Psi(\vz)$ close to the assumed latent prior distribution $\Pi(\vz)$ (\cite{dai2019diagnosing,hoffman2016elbo,burgess2018understanding}). This can be achieved either by (i) assuming a pre-defined primitive distribution for $\Pi(\vz)$ and modifying the Encoder such that $\Psi(\vz)$ follows assumed $\Pi(\vz)$ (\cite{kingma2013autoencoding,44904,46498,chen2018isolating,higgins2017beta,kim2018disentangling,kingma2016improved}) or by (ii) modifying the latent prior $\Pi(\vz)$ to follow whatever distribution $\big(\Psi(\vz)\big)$ Encoder imposes on the latent space (\cite{tomczak2017vae,bauer2018resampled,klushyn2019learning,hoshen2019non,van2017neural}). 
%We briefly review some of the aforementioned methods in the next section and subsequently provide motivation for the current work. 
\par
The seminal paper on VAE (\cite{kingma2013autoencoding}) proposes a probabilistic Encoder which is tuned to output the parameters of the conditional posterior $\Psi(\vz|\vx)$ which is forced to follow the Normal distribution prior assumed on $\Pi(\vz)$. However, the minimization of the divergence between the conditional latent distribution and the prior in the VAE leads to trade-off between the reconstruction quality and the latent matching, as this procedure also leads to the minimization of the mutual information between $\mathcal{X}$ and $\mathcal{Z}$, which in turn reduces Decoder's ability to render good reconstructions (\cite{kim2018disentangling}). This issue is partially mitigated by altering the weights on the two terms of the ELBO during optimization (\cite{higgins2017beta,burgess2018understanding}), or through introducing explicit penalty terms in the ELBO to strongly penalize the deviation of $\Psi(\vz)$ from assumed prior $\Pi(\vz)$ (\cite{chen2018isolating,kim2018disentangling}). Adversarial Auto-Encoders (AAE) (\cite{44904}) and Wasserstein Auto-Encoders (WAE) (\cite{46498}) address this issue, by taking advantage of adversarial training to minimize the divergence between $\Psi(\vz)$ and $\Pi(\vz)$, via deterministic Encoder and Decoder networks. There also have been attempts in employing the idea of normalizing flow for distributional estimation for making $\Psi(\vz)$ close to $\Pi(\vz)$ (\cite{kingma2016improved,rezende2015variational}). These methods, although improve the generation quality over vanilla VAE while providing additional properties such as disentanglement in the learned space, fail to match the generation quality of GAN and its variants.

In another class of methods, the latent prior $\Pi(\vz)$ is made learnable instead of being fixed to a primitive distribution so that it matches with Encoder imposed $\Psi(\vz)$. In VamPrior (\cite{tomczak2017vae}), the prior is taken as a mixture density whose components are learned using pseudo-inputs to the Encoder. \cite{klushyn2019learning} introduces a graph-based interpolation method to learn the prior in a hierarchical way. In \cite{van2017neural, kyatham2019variational}, discrete latent space is employed, using vector quantization schemes where the prior is learned using a discrete auto-regressive model. \textcolor{blue}{While these prior matching methods provide various advantages, there is no mechanism to ward-off the `spurious' latent dimensions that are known to degrade the generation quality. While there exists a possibility that the Decoder learns to ignore those spurious dimensions by making the corresponding weights zero there is no guarantee or empirical evidence of neglecting those dimensions. Another indirect approach to handle this issue might be adding noise to the input data. However, this approach avoids the problem instead of solving it. To summarize, it is observed that, without additional modifications, in vanilla AE-based models, the existence of superfluous latent dimensions degrade the generation quality (\cite{dai2019diagnosing}). Motivated by the aforementioned observations, ours is the first work that explicitly looks at the effect of latent dimensions on the generation quality of AE-based models. Further, unlike previous works, we attempt to solve this issue explicitly, instead of relying on decoder statistics or noise-based heuristics.}

%% file: theory.tex
\section{\MakeUppercase{Effect of Latent Dimensionality}}\label{sec:theory}

\subsection{PRELIMINARIES}\label{3.1}
In this section, we theoretically examine the effect of latent dimensionality on the quality of generated data in AE based generative models.  We show that if dimensionality of the latent space $\mathcal{Z}$ is more than the \textit{optimal} dimensionality (to be defined), $\Pi(\vz)$ and $\Psi(\vz)$ diverge too much whereas it being less leads to information loss.
\begin{figure}[ht]
    \begin{center}
         \includegraphics[clip, trim=0.0cm 0.1cm 0.0cm 0.11cm, width=\linewidth, keepaspectratio]{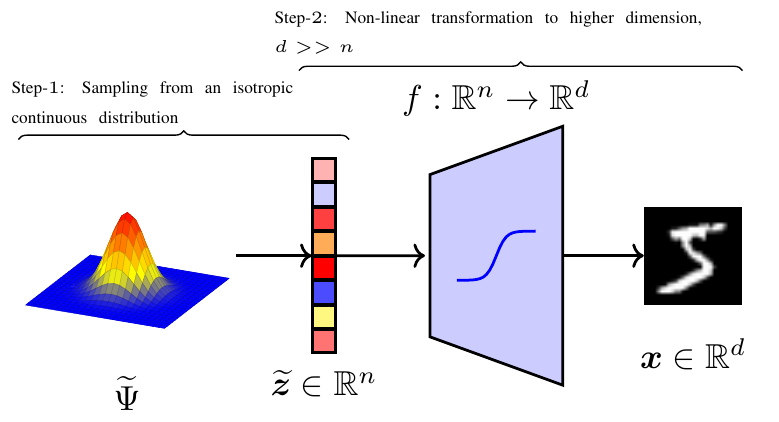}
    \end{center}
    \caption{Depiction of the assumed data generation process. Samples drawn from a `true' latent distribution  $\widetilde\Psi(\widetilde{\mathbf{z}})$ are passed through a function $f$ to obtain $\vx$.}
    \label{fig:dataGenerationHypothesis}
\end{figure}
\par To start with, we allow a certain inductive bias in assuming that nature generates the data as described in Figure \ref{fig:dataGenerationHypothesis} using the following two-step process: First sample from some isotropic continuous latent distribution in $n$-dimensions (call this $\widetilde\Psi$ over $\widetilde{\mathcal{Z}}$), and then pass this through a function $f:\real{n}\to\real{d}$, where $d$ is the dataset dimensionality. Typically $d>>n$, thereby making data to lie on a low-dimensional manifold in $R^d$. Since $\widetilde{\mathcal{Z}}$ can intuitively be viewed as the latent space from which the nature is generating the data, we call $n$ the \textit{true latent dimension} and function $f$, as the \textit{data-generating function}. Note that within this ambit,  $\widetilde{\mathcal{Z}}$ forms the domain of $f$ and it is unique only up to its range with the following properties:

\begin{enumerate}
    \item[A1] $f$ is $L$\textit{-lipschitz}: $\exists$ some finite $L\in\real{+}$ satisfying $||f(\widetilde\vz_1)-f(\widetilde\vz_2)|| \le L|| \widetilde\vz_1-\widetilde\vz_2||,\,\,\forall \widetilde\vz_1,\widetilde\vz_2\in \widetilde{\mathcal{Z}}$.
    \item[A2]There does not exist $f^*:\real{n'}\to\real{d},n'<n$ satisfying A1 such that the range of $f$ is a subset of the range of $f^*$. 
    \end{enumerate}
%, while ReLU activations satisfy the relaxed version of the condition
%({\color{teal} explain relaxed?})({\color{red} LHS$\leq$ RHS is what we really need, the equality is never used in the proof. In fact, L-lipshitz does not require the equality, as far as I know.}) {\color{teal} Precisely. I have changed it, we do not limiting argument.}
\textcolor{blue}{The first property is satisfied by a large class of functions, including neural networks and the second simply states that $n$, the dimension of the domain (generative latent space) of $f$ is minimal\footnote{If there exists such an $f^*$, then that would become the generating function with $n'$ being minimal.}. Hence, it is reasonable to impose these restrictions on data-generating functions. (An illustrative example is provided in the supplementary material.)} %in the absence which, the choice of $n$ becomes ambiguous, since given any $f$, one an always define $f'$ in a higher dimension 
%which has the exact same Range as $f$.}\\ 
%Intuitively this means that each of the latent dimension contribute in generation of some small (albeit infinitesimal) region in the domain of the observed data but not that every latent dimension affect every observed data point.} %in the domain\footnote{In mathematical terms, this is equivalent to requiring a full rank Jacobian, which is a strictly stronger assumption}. For example, consider the case where a leaf is being photographed. A young leaf in broad daylight has colour roughly (120,100,50) in the HSL system. As the age of the leaf increases, the lightness starts to fall, but a similar fall in lightness will also be observed with fading daylight. At this point, lighting conditions and age of leaf have identical effect on the appearance of the leaf. However, after a point, age will start reducing the hue of the leaf, while lighting conditions will continue decreasing its lightness. Since at this point these two factors influence the outcome differently, they can be separate factors in our input space. On the other hand, the distance from which the photo was taken and the optical zoom of the lens will always have similar effect, and therefore, only one of these is allowed as a factor. Note that this example is presented for illustrative purposes only, and in real cases, the input factors are unlikely to directly map to real-world causes.}

\subsection{CONDITIONS FOR GOOD GENERATION}\label{3.2}
In this section, we formulate the conditions required for faithful generation in latent variable generative models.  Let $ \Gamma(\vx,\vz)$ and $\Gamma'(\vx,\vz)$ denote the true and the (implicitly) inferred joint distribution of the observed and latent variables. The goal of latent variable generative models is to minimize the negative log-likelihood of $\Gamma'(\vx,\vz)$ under $ \Gamma(\vx,\vz)$:
%The goal of any generative model is to minimise the divergence between the generated distribution $\Upsilon'(\vx)$ and the true data distribution $\Upsilon(\vx)$. This objective is typically expressed as minimising the KL-divergence between the two distributions. Since auto-encoder based frameworks work through a latent space $\mathcal{Z}$, their objective is to minimise the divergence between the joint distribution of $\vx$ and $\vz$. This can be simplified as follows:

\begin{equation}
\mathcal{L}(\Gamma,\Gamma') =-\mathop{\mathbb{E}}_{\vx,\vz\sim\Gamma}\big[\log(\Gamma'(\vx,\vz))\big]
\label{lh}
\end{equation}

%Since the entropy of the true joint distribution $ \Gamma(\vx,\vz)$  cannot be controlled, the objective in Eq. \ref{lh} can be equivalently cast as minimizing the KL-divergence between $\Gamma'(\vx,\vz)$ and $ \Gamma(\vx,\vz)$:

% %\begin{align}
% \begin{split}
% D_{KL}(\Gamma(\vx,\vz)||\Gamma'(\vx,\vz))&=\mathop{\mathbb{E}}_{\vx,\vz\sim\Gamma}(\log\frac{\Gamma(\vx,\vz)}{\Gamma'(\vx,\vz)})\\ &=\mathop{\mathbb{E}}_{\vx,\vz\sim\Gamma}(-\log(\Gamma'(\vx|\vz)))\\
% &\quad+\mathop{\mathbb{E}}_{\vx,\vz\sim\Gamma}(\log\frac{\Gamma(\vx,\vz)}{\Gamma'(\vz)})
% \end{split}
% \end{align}
%Since $\Upsilon$ and $\Upsilon'$ were originally defined over only $\vx$, 
% Since we are free to choose $\Gamma'(z)$, We set this to our chosen prior $\Pi(\vz)$. 
% Further, since AAEs use deterministic functions to map from $\vx$ to $\vz$ and back, we replace $\Gamma$ in the second term with $\Psi$\footnote{formally, this term splits into $\Psi(\vz)\Gamma(\vx|\vz)$, of which the second term is 1 over its entire support, thereby making its logarithm $0$}, as defined in Section \ref{sec:related}. 
An AE-based generative model would attempt to minimize Eq. \ref{lh} by learning two parametric functions,  $E_{\kappa}\triangleq g:\real{d}\to\real{m}$ ($m$ is hereafter referred to as \textit{assumed latent dimension / model capacity}) and $D_{\psi}\triangleq g':\real{m}\to\real{d}$, to approximate the distributions $\Psi(\vz|\vx)$ and $\Gamma(\vx|\vz)$, respectively. Further, Eq. \ref{lh} can be broken down into two terms, and the objective of any AE based model can be restated as:
\begin{equation}
    \min\bigg(\underbrace{\mathop{\mathbb{E}}_{\Gamma}[-\log(\Gamma'(\vx|\vz))]}_\text{R1}+\underbrace{\mathop{\mathbb{E}}_{\Gamma}[\log\frac{1}{\Gamma'(\vz)}]}_\text{R2}\bigg) 
    \label{elbo}
\end{equation}
If $E_{\kappa}$ and $D_{\psi}$ are deterministic (as in the case of AAE (\cite{44904}), WAE (\cite{zhang2019wassersteinwasserstein}) etc.), then the two terms in Eq. \ref{elbo} can be cast as the following two requirements (see the supplement for the proof):

\begin{enumerate}
\item[R1] $ f(\widetilde\vz)=g'(g(f(\widetilde\vz)))\;\forall\  \widetilde\vz \ \in \real{n}$. This condition states that the reconstruction error between the real and generated data should be minimal.

    %\item[R2] $\not\exists h:\real{m}\to\{0,1\}$, such that $\int_{\real{m}}h(\vx})d\vx=0$ and $h(g(f(\vx)))=1\;\forall \vx\in \real{n}$.{\color{teal} this requirement is not clear? Is it ever used in the argument below?}{\color{red} basically that range of g(f(x)) has measure 0 in its output space. Also, this h is exactly what the discriminator needs to learn in order to win the adverserial game. It is used in the second last paragraph.} ({\color{teal} I see but we have to motivate this assumption in a different way because we are not using any reference to AAE here, this is a general section about AE and that we need matching of dimensions})
    \item[R2] The Cross Entropy $\mathcal{H}(\Psi,\Pi)$ between the chosen prior $\Psi$, and $\Pi$ on $\mathcal{Z}$ is minimal.
\end{enumerate}
With this, we state and prove the conditions required to ensure R1 and R2 are met with assumed data generation process. 

\begin{theorem} With the assumption of data generating process mentioned in Sec.\ref{3.1}, requirements R1 and R2 (Sec.\ref{3.2}), can be satisfied iff \textit{assumed latent dimension} $m$ is equal to \textit{true latent dimension} $n$.
\end{theorem}
\noindent
\textbf{Proof:} We prove by contradicting either R1 or R2, in assuming both the cases of $m<n$ or $m>n$.
\\\\\noindent
% \textit{Case A : $m<n$} : Since $f$ is injective and differentiable everywhere, it must have a continuous left inverse (call it $f^{-1}$). Also, the R1 forces $g'$ to be the left inverse of $g$ on the range of $f$. Since $g$ and $g'$ are both neural networks, the composite $g\circ f$ has a continuous left inverse $f^{-1}\circ g'$ which is impossible due to the \vspace{2mm}following lemma: 
% \\
% \noindent
% \textbf{Lemma 1}: A continuous function $\displaystyle\phi:\real{\alpha}\to\real{\beta}$ cannot have a continuous left inverse if $\alpha>\beta$. (Proof is given in the Supplementary \vspace{2mm} material).
% \textbf{Proof}: It follows trivially from the fact that such a function would define a homeomorphism from $\real{\alpha}$ to a subset of  $\real{\beta}$, whereas it is well known that these two spaces are not homeomorphic.\qed
% \\ This implies that R1 and Lemma 1 contradict each other in the case $m<n$ and thus to obtain a good reconstruction, $m$ should at-least be equal to $n$.
%I can add a proof, but it seems to be a digression%
%\\\\In a more intuitive sense, we cannot compress the information present in n dimensions into n dimensions. Therefore, we cannot satisfy the first condition if we have $m<n$\\\\
\noindent
\textit{Case A $(m < n)$:} For R1 to hold, the range of $f$ must be a subset of the range of $g'$. Further, since $g'$ is a Neural Network, it satisfies A1. But, by A2, such a function cannot exist if $m<n$.\\
\textit{Case B $(m>n)$}: %The intuition here is that if we map a smaller dimensional space to a larger dimensional space, we would be able to cover only a negligible fraction of the higher dimensional space. \\\\
For the sake of simplicity, let us assume that $\widetilde{\mathcal{Z}}$ is a unit cube\footnote {One can easily obtain another function $\nu:[0,1]^{n}\to \widetilde{\mathcal{Z}}$ that scales and translates the unit cube appropriately. Note that for such a $\nu$ to exist, we need $\widetilde{\mathcal{Z}}$ to be bounded, which may not be the case for certain distributions like the Gaussian distributions. Such distributions, however, can be approximated successively in the limiting sense by truncating at some large value \cite{rudin1964principles}} in $\real{n}$. We show in Lemma 2 and 3 that in this case, R2 will be contradicted if $m>n$. The idea is to first show that the range of $g\circ f$ will have Lebesgue measure 0 (Lemma 1) and this leads to arbitrarily large $\mathcal{H}$ (Lemma 2).   \\\ %Since $f,\ g$ are Lipschitz $g\circ f$ must have a range with Lebesgue measure 0 as a consequence of the following lemma:\\\\
\noindent
\textbf{Lemma 1:}
Let $\Omega:[0,1]^{\alpha}\to \real{\beta}$ be an $L-lipschitz$ function. Then its range $R\in\real{\beta}$ has Lebesgue measure $0$ in $\real{\beta}$ dimensions if $\beta>\alpha$.
\\\\
\textbf{Proof:}
\noindent\vspace{2mm}
For some $\epsilon\in\mathbb{N}$, consider the set of points:\\
\vspace{2mm}
\noindent
S={\large\{}$(\frac{a_0+0.5}{\epsilon},\ldots,\frac{a_{\alpha-1}+0.5}{\epsilon})\big|a_i\in\{0,\ldots,\epsilon-1\}${\large \}}.\\
Construct closed balls around them having radius $\frac{\sqrt{\alpha}}{2\epsilon}$. It is easy to see that every point in the domain of $\Omega$ is contained in at least one of these balls. This is because, for any given point, the nearest point in S can be at-most $\frac{1}{2\epsilon}$ units away along each dimension. Also, since $\Omega$ is $L$-lipschitz, we can conclude that the image set of a closed ball having radius $r$ and centre $\vu\in[0,1]^\alpha$ would be a subset of the closed ball having centre $\Omega(\vu)$ and radius $L \times r$. \\
The range of $\Omega$ is then a subset of the union of the image sets off all the closed balls defined around S. The volume of this set is upper bounded by the sum of the volumes of the individual image balls, each having volume $\frac{c}{\epsilon^\beta}$  where c is a constant having value $\frac{(L)^\beta(\alpha\pi)^{\frac{\beta}{2}}}{\Gamma(\frac{\beta}{2}+1)}$.
Therefore,  
\begin{equation}\label{goestozero}
\begin{split}
    \mathrm{vol(R)}&\leq
|S|\times\frac{c}{\epsilon^\beta}
=\frac{c}{\epsilon^{\beta-\alpha}}.
\end{split}
\end{equation}
The final quantity of Eq. \ref{goestozero} can be made arbitrarily small by choosing $\epsilon$ appropriately. Since the Lebesgue measure of a closed ball is same as its volume, the range of $\Omega$, $R$ has measure $0$ in $\real{\beta}.$\qed
% \textbf{Note:} While the condition {\color{teal} you mean Lipschitznes} on the function might look complex, it is only saying that all partial derivatives exist and are finite at all points, i.e. the function is differentiable everywhere in its domain. In fact, we could relax this condition and allow the function to be non-differentiable at some points, as long as it is continuous and all directional derivatives are bounded. In such a case, the equality will change into an inequality(i.e. every limit is less than L), but the proof still works, as it never uses the equality.{\color{red} This note is kind of redundant now that you have talked about this in the f is L-lipshitz point}{\color{teal} I agree you can strike this off}.\\
\\Since $f,\text{ and } g$ are Lipschitz, $g\circ f$ must have a range with Lebesgue measure 0 as a consequence of Lemma 2. Now we show that as a consequence of the range of $g\circ f$ (call it $\mathcal{R}$) having measure $0$, the cross-entropy between $\Pi$ and $\Psi$ goes to \vspace{2mm}infinity.\\
\textbf{Lemma 2:} If $\Pi$ and $\Psi$ are two distributions as defined in Sec.\ref{3.1} such that the support of the latter has a $0$ Lebesgue measure, then  $\mathcal{H}(\Pi,\Psi)$ grows to be arbitrarily large.\\
\textbf{Proof}: $\Psi $ can be equivalently expressed as:
\begin{equation}\label{redef}
    \Psi(\vz)=\begin{cases}
    \widetilde\Psi(\widetilde\vz) & \text{if }\exists\ \widetilde\vz\text{\footnotemark}\in\widetilde{\mathcal{Z}} \text{ s.t. }g(f(\widetilde\vz))=\vz,\\
    0 & \text{otherwise}
    \end{cases}
\end{equation}
\footnotetext{Note that in general, $\widetilde\vz$ is not unique, and if multiple such $\widetilde\vz$ exist, we have to sum(or perhaps integrate) $\widetilde\Psi$ over all such $\widetilde\vz$}
 Define $\mathbb{I}_{\mathcal{R}}$ as the indicator function of $\mathcal{R}$, i.e. 
\begin{equation}
    \mathbb{I}_{\mathcal{R}}(\vz)=\begin{cases}
    1 & \text{if } \exists~\widetilde\vz\in\widetilde{\mathcal{Z}} \text{ s.t. }g(f(\widetilde\vz))=\vz,\\
    0 & \text{otherwise}
    \end{cases}
\end{equation}
Since $\mathcal{R}$ has measure $0$ (Lemma 2), we have
\begin{equation}\int_{\real{m}}  \mathbb{I}_{\mathcal{R}}({\vz})d\vz=0\end{equation}
\noindent
Further, since $\mathbb{I}_{\mathcal{R}}$ is identically $1$ in the support of  $\Psi$, we have
\begin{equation}
    \Psi(\vz)=\Psi(\vz)\mathbb{I}_{\mathcal{R}}(\vz)
\end{equation}
\noindent
Now consider the cross-entropy between $\Pi$ and $\Psi$ given by:
\begin{equation}\label{KL}
\begin{split}
    \mathcal{H}(\Pi,\Psi)&=\int_{\mathcal{Z}}\Pi(\vz)(-\log(\Psi(\vz)))d\vz\\
    &\geq\int_{\mathcal{Z}-\mathcal{R}}\Pi(\vz)(-\log(\Psi(\vz)\mathbb{I}_{\mathcal{R}}(\vz)))d\vz\\
    &\geq \varrho\int_{\mathcal{Z}-\mathcal{R}}\Pi(\vz)d\vz\;
\end{split}
\end{equation}
for any arbitrarily large positive real $\varrho$. This holds true because $\mathbb{I}_{\mathcal{R}}$ is identically 0 over the domain of integration. Further,
\begin{equation}\label{prob}
    \begin{split}
        \int_{\mathcal{Z}-\mathcal{R}}\Pi(\vz)&\geq\int_{\mathcal{Z}}\Pi(\vz)-\int_{\mathcal{R}}\Pi(\vz)\\
        &=1-\int_{\real{m}} \Pi(\vz)\mathbb{I}_{\mathcal{R}}({\vz})d\vz\\
        &\geq 1- \max_{\real{m}}(\Pi(\vz))\int_{\real{m}} \mathbb{I}_{\mathcal{R}}({\vz})d\vz\\
        &=1
    \end{split}
\end{equation}
Combining \ref{KL} and \ref{prob}, the required cross-entropy is lower bounded by an arbitrarily large quantity $\varrho$.\qed \\\
Thus Lemma 2 contradicts R2 required for good generation when $m>n$. Therefore, to ensure good generation neither $m>n$ nor $m<n$ can be true. Thus, the only possibility is $m=n$. This concludes Theorem 1.\qed \\
\\\\
One can ensure good generation, by satisfying both R1 and R2 via a trivial solution in the form of $g'=f$ with an appropriate $g$ and making $m=n$. However, since neither $n$ nor $f$ is known, one needs a practical method to ensure $m$ to approach $n$ which is described in the next section. 

% {\color{red}\textit{Case B : $m < n$:} From R1, we have
% \begin{equation}
%     f(\widetilde\vz)=g'(g(f(\widetilde\vz)))\;\forall\  \widetilde\vz \ \in \real{n}
% \end{equation}
% passing both sides through $f^*$, as defined in A1, gives us
% \begin{equation}
%     f^*(f(\widetilde\vz))=f^*(g'(g(f(\widetilde\vz))))\;\forall\  \widetilde\vz \ \in \real{n}
% \end{equation}
% The composite function on RHS has a range which is the subset of the range of $f^*\circ z$, which by Lemma 2, is known to has measure 0 in $\real{n}$. The composite function on LHS, however, has range having measure non-zero in $\real{n}$. This gives us a contradiction, and therefore, we cannot have $m<n$.} \\

%% file: model.tex
\section{MaskAAE (MAAE)}\label{sec:model}

\subsection{\MakeUppercase{Model description}}

Our premise in section \ref{3.2} demands a pair of deterministic Encoder and Decoder networks satisfying R1 and R2, to ensure good quality generation. AE-models with deterministic \encoder and \decoder networks, such as Adversarial Auto-Encoder (AAE) (\cite{44904}) and Wasserstein Auto-Encoder (WAE) (\cite{zhang2019wassersteinwasserstein}) implement R1 by approximating norm-based losses and R2 through an adversarial training mechanism under metrics such as JS-Divergence or Wasserstein distance. However, most of the time, the choice of the latent dimensionality is ad hoc and there is no mechanism to get rid of the excess latent dimensions that is critical for good-quality generation as demanded by Theorem 1. Therefore, in this section, we take the ideas presented in Section~\ref{sec:theory}, and propose an architectural modification on models such as AAE/WAE, such that being initialized with a large enough estimated latent space dimension the model would learn a binary-mask automatically discovering the right number of latent dimensions required.

%further establishes that the requirements are satisfied iff assumed latent dimension $m$ is equal to actual latent dimension $n$. %As pointed out by \cite{dai2019diagnosing}, a single-stage VAE resolves this issue partially by learning a ``parsimonious representation of low-dimensional manifolds''. However, a single-stage VAE does not necessarily learn the correct probability measure within such a manifold. As a remedy to this problem, \cite{dai2019diagnosing} introduced another VAE network that learns to sample from the latent distribution of the first stage. 
\begin{figure}[ht]
    \begin{center}
         \includegraphics[width=\linewidth, keepaspectratio]{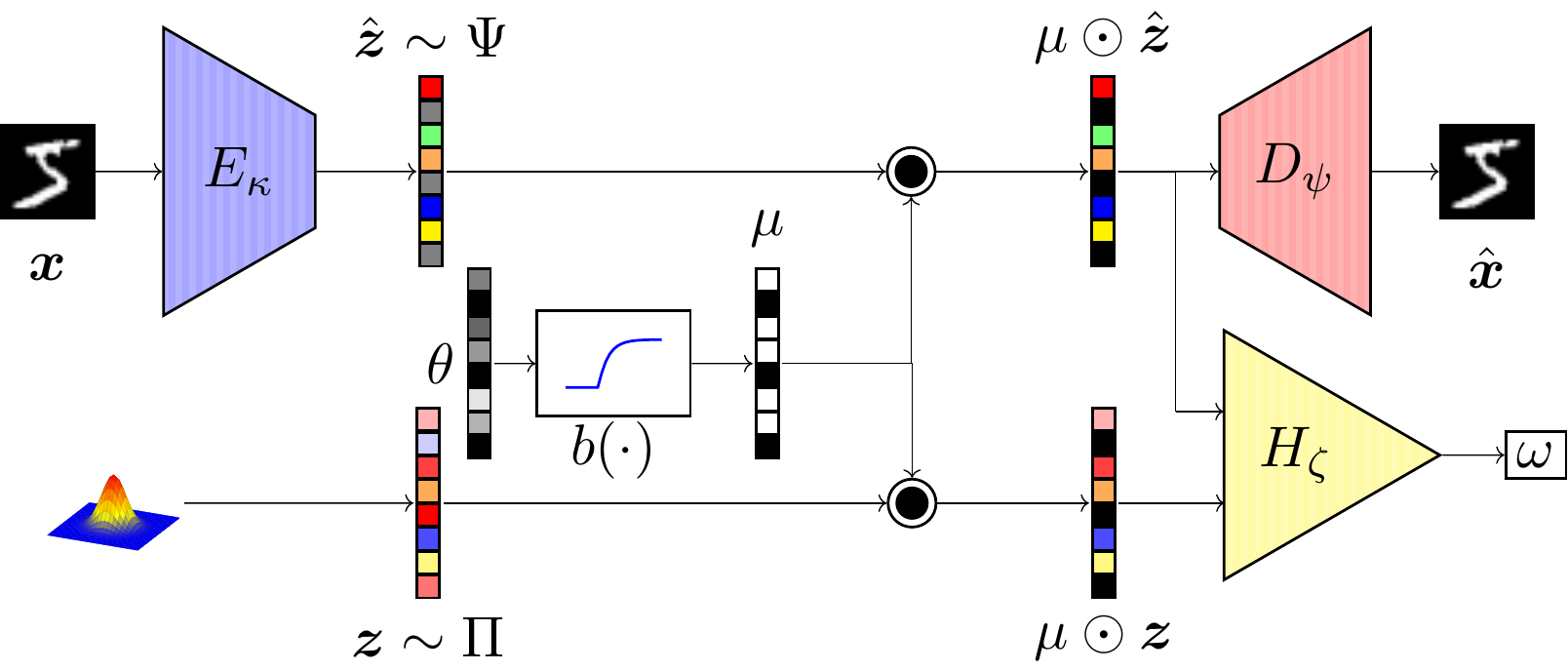}
    \end{center}
    \vspace{-2mm}
    \caption{Block Diagram of MaskAAE. It consists of an encoder, $E_\kappa$, a decoder, $D_\psi$, and a discriminator $H_\zeta$ as in AAE. A new layer called mask, $\mu$ is introduced at the end of the encoder to suppress spurious latent dimensions. The prior also gets multiplied with the same mask before going into the Discriminator to ensure prior matching (R2).}
    \vspace{-4mm}
    \label{fig:architecture}
\end{figure}

 %Implementation wise two aspects require attention.
%\begin{itemize}
  %  \item The decoder network should be able to neglect the excess dimensions in the latent representation while reconstructing or generating images.
  %  \item The encoder network should not map the input data to the excess dimensions in the latent representation. In other words, excess dimensions in the latent representation should have zero variance so that those dimensions contain no information about input data.
%\end{itemize}{}

Specifically, we propose the following modifications in the  AAE-like architecture (\cite{44904,zhang2019wassersteinwasserstein}), which contain an additional component called Discriminator (\discriminator) that is used to match $\Psi(\vz)$ and $\Pi(\vz)$ via adversarial learning. Our model, called the MaskAAE is detailed in figure \ref{fig:architecture}. 
\begin{enumerate}

\item We introduce a trainable mask layer, $\mu \in \{0, 1\}^m$, just after the final layer of the Encoder network. 
\item Before passing the encoded representation, $\hat\vz$ of an input image $\vx$ to the decoder network (\decoder) and the Discriminator network (\discriminator) a Hadamard product is performed between $\hat\vz$ and $\mu$.
\item A Hadamard product is performed between the prior sample, $\vz \sim \Pi(\vz)$ and the same mask $\mu$ as in item (1), before passing it as an input to the discriminator network \discriminator to ensure R2. 
\item During inference, the prior samples are multiplied with the learned mask before giving as input to the  Decoder (\decoder) network which serves the generator. 
\end{enumerate}
Intuitively, masking of both the encoded latent vector and prior with a same binary mask allows us to work only with a subset of dimensions in the latent space. This means that even though $m$ (the initial assumed latent dimensionality) may be greater than $n$, mask (if learned properly) reduces the  encoded latent space to $\mathcal{R}^n$. This will in-turn facilitate better matching of $\Psi(\vz)$ and $\Pi(\vz)$  (R2) required for better generation. 

\subsection{{TRAINING MaskAAE}}

MaskAAE is trained exactly similarly as one would train an AAE/WAE but with the addition of a loss term to train the mask layer. Here, we provide the details of the mask-loss only. For a complete description of other AAE/WAE based training loss terms refer to the supplementary material. 

Although, the mask by definition is a binary-valued vector, to facilitate gradient flow during training, we relax it to be continuous valued while penalizing it for deviation from either $0$ or $1$. Specifically, we parameterize $\mu$ using a vector $\theta \in \real{m}$ such that $\mu=b(\theta)$ where, $b(\theta)=\max(0,1-e^{-\theta})$. $\theta$ is initialized by drawing samples from $\mathcal{U}[0,a]$, where $a\in\mathbb{Z}^+$. Intuitively, this parameterization bounds $\mu$ in the range $(0,1)$. Since the mask layer affects both the requirements R1 and R2, it is trained so as to minimize both the norm-based reconstruction error (first term in Eq. \ref{maskloss}) and divergence metrics such as JS-divergence or Wasserstein's distance, between the masked prior distribution and the masked encoded latent distribution (second term in Eq. \ref{maskloss}). Finally, a polynomial regularizer (third term in Eq. \ref{maskloss}) is also added on $\mu$ so that any deviation from $\{0,1\}$ is penalized.  Therefore, the final objective function for the mask layer, $L_{mask}$ consists of three terms as below.
        \begin{equation}
        \label{maskloss}
        \begin{split}
            L_{mask} &= \frac{\lambda_1}{s}\sum_{i=1}^{s}||\vx^{(i)}-D_\psi(\mu \odot E_\kappa(\vx^{(i)}))|| \\
            &\qquad+ \lambda_2(1 + \omega)^2 
            %&\qquad
            + \lambda_3 \sum_{j=1}^{m}|\mu_j(\mu_j - 1)|
        \end{split}\end{equation}
        where, 
$\omega = \frac{1}{s}\sum_{i}H_\zeta(\mu \odot \vz^{(i)}) - \frac{1}{s}\sum_{i}H_\zeta(\mu \odot E_\kappa (\vx^{(i)}))$ is the Wasserstein's distance, $s$ denotes batch size, and the weights $(\lambda_1$, $\lambda_2$, $\lambda_3)$ of different loss terms are hyper-parameters. % \par
%We use norm based loss functions (mean square error, mean absolute error) or binary cross entropy for minimizing the reconstruction error. However, a new term, variance penalty, is added to the standard AAE reconstruction objective to penalize non-zero variance across the spurious dimensions, i.e., dimensions corresponding to minimal(close to zero) mask value.
Details about the training algorithm, and the architectures for \encoder, \decoder and \discriminator are available in the supplementary material.

%% file: expt.tex
\section{EXPERIMENTS AND RESULTS}\label{sec:expt}
We divide our experiments into two parts: (a) Synthetic, and (b) Real. In synthetic experiments, we control the data generation process, with a known number of true latent dimensions. Hence, we can compare the performance of our proposed model for several true latent dimensions, and examine whether our method can discover the true number of latent dimensions. This also helps us validate some of the theoretical claims made in Section~\ref{sec:theory}. On the other hand, the objective of the experiments with real datasets is to examine whether our masking based approach can result in a better generation quality as compared to the state-of-the-art AE-based models. We would also like to understand the behaviour of the number of dimensions which are masked in this case (though the precise number of latent data dimensions may not be known).

\begin{figure*}
    \begin{center}
        \includegraphics[clip, trim=0.0cm 0.0cm 0.0cm 0.0cm, width=\linewidth, keepaspectratio]{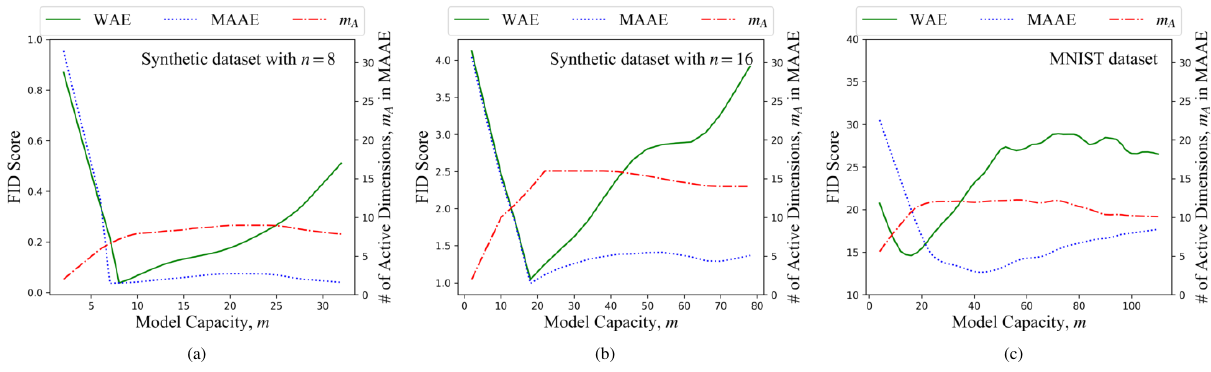}
    \end{center}
    \vspace{-4mm}
    \caption{(a) and (b) shows FID score for WAE and MAAE and active dimension in a trained MAAE model with varying model capacity, $m$  for synthetic dataset of true latent dimensions, $n=8$  and $n=16$, $m_{A}$ represents the number of unmasked latent dimensions in the trained model and (c) shows the same plots for MNIST dataset.}
    \vspace{-2mm}
    \label{fig:synthetic}
\end{figure*}

\subsection{SYNTHETIC EXPERIMENTS}
In the following description, we will use $n$ to denote the true latent dimension, and $m$ to denote the assumed latent dimension (or model capacity) in line with the notation used earlier in the paper. Assuming that data is generated according to the generation process described in Section~\ref{sec:theory}, we are interested in answering the following questions: (a) Given sufficient model capacity (i.e, $m \ge n$ and sufficiently powerful $E_\kappa$, $D_\psi$ and $H_\zeta$), can MAAE discover the true number of latent dimensions? (b) What is the quality of the data generated by MAAE for varying values of $m$?

In an ideal scenario, we would expect that whenever $m \ge n$, MAAE masks $(m-n)$ number of dimensions. Further, we would expect that the performance of MAAE is independent of the value of $m$, whenever $m \ge n$. For each value of $m$ that we experimented with, we also trained an equivalent WAE model with exactly the same architecture for \encoder, \decoder and \discriminator as in MAAE without the mask layer. We would expect the performance of the WAE model to deteriorate in cases whenever $m < n$ or $m > n$ if our theory were to hold correct.
 
 In line with our assumed data generation process, the data for our synthetic experiments is generated using the following process.
\begin{itemize}
    \item Sample $\tilde{\vz} \sim \mathcal{N}(\mu_s,\Sigma_s)$, where the mean $\mu_s \in \real{n}$ was fixed to be zero and $\Sigma_s \in \real{n \times n}$ represents the diagonal co-variance matrix (isotropic Gaussian).
    \item Compute $\vx=f(\tilde{\vz})$, where $f$ is a non-linear function computed using a two-layer fully connected neural network with $k$ units in each layer, $d>>n$ output units, and using leaky ReLU as the non-linearity (refer to the supplement for more details). The weights of these networks are randomly fixed and $k$ was taken as 128. 
\end{itemize}
We set $n=8$ and $16$, and varied $m$ in the range of $[2,32]$ and $[2,78]$ with step size $2$, for $n=8$ and $n=16$ respectively. We use the standard Fr\'echet Inception Distance (FID) (\cite{NIPS2017_7240}) score between generated and real images to validate the quality of the generated data, because FID has been shown to correlate well the human visual perception and also sensitive to artifacts such as mode collapse (\cite{lucic2018gans, NIPS2018_7769}). Figure~\ref{fig:synthetic} (a) and (b) presents our results on synthetic data. On X-axis, we plot $m$ and Y-axis (left) plots the FID score comparing MAAE and WAE for different values of $m$. Y-axis (right) plots the number of active dimensions discovered by our algorithm.
It is seen that both MAAE and WAE, achieve the best FID score when $m=n$. But whereas the performance for WAE deteriorates with increasing $m$, MAAE retains the optimal FID score independent of the value of $m$. Further, in each case, we get very close to the true number of latent dimensions, even with different values of $m$ (as long as $m > 8$ or $16$, respectively). This clearly validates our theoretical claims, and also the fact that MAAE is capable of offering good quality generation in practice.

% This clearly validates our theoretical claims, and also the fact that our model is effective in practice in offering good quality generation. 

\begin{figure*}[ht]
    \begin{center}
        \includegraphics[clip, trim=0.0cm 0.1cm 0.0cm 0.01cm, width=\linewidth, keepaspectratio]{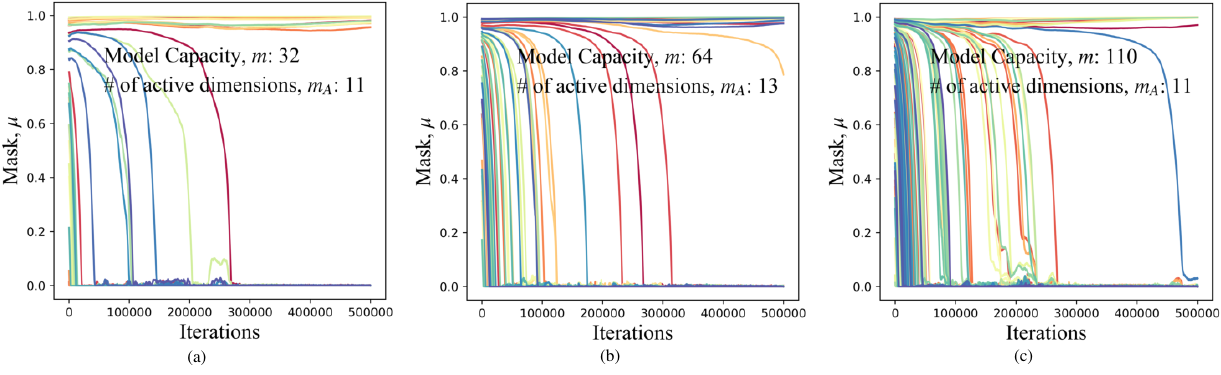}
    \end{center}
    \vspace{-2mm}
    \caption{Behaviour of mask in MAAE models with different $m$ for the MNIST dataset. Model capacity, $m$, in figure (a), (b), and (c) are $32,~ 64,$ and $110$, respectively. The active dimensions after training are $m_A$ are $11,~ 13,$ and $11$ respectively.}
    \vspace{-4mm}
    \label{fig:maskBehaviour}
\end{figure*}

\subsection{REAL EXPERIMENTS}
In this section, we examine the behavior of MAAE on real-world datasets. In this case, the true latent data dimensions ($n$) is unknown, but we can still analyze the behavior as the estimated latent number of dimension ($m$) is varied. We work on four image datasets used extensively in the literature:
(a) MNIST (\cite{MNIST}) (b) Fashion MNIST (\cite{xiao2017fashionmnist}) (c) CIFAR-10 (\cite{Krizhevsky09learningmultiple}) (d) CelebA (\cite{liu2015faceattributes}) with standard test/train splits.

In our first set of experiments, we perform an analysis similar to the one done in the case of synthetic data, for the MNIST dataset. Specifically, we varied the estimated latent dimension (model capacity $m$) for MNIST from $10$ to $110$, and analyzed the FID score, as well as the true dimensionality as discovered by the model. For comparison, we also did the same experiment using the WAE model. Figure~\ref{fig:synthetic} (c) shows the results. As in the case of synthetic data, we observe a U-shape behavior for the WAE model, with the lowest value achieved at $m=13$. This validates our thesis that the best performance is achieved at a specific value of latent dimension, which is around $13$ in this case. Further, looking at MAAE curve, we notice that the performance (FID score) more or less stabilizes for values of $m \ge 10$. In addition, the true latent dimension discovered also stabilizes around $10-13$ irrespective of $m$, without compromising much on the generation quality. Note that the same network architecture was used at all points of Figure~\ref{fig:synthetic}. These observations are in line with the expected behavior of our model, and the fact that our model can indeed mask the spurious set of dimensions to achieve good generation quality.

Figure \ref{fig:maskBehaviour} shows the behaviour of mask for model capacity  $m = 32, 64$ and $110$ on MNIST dataset. Interestingly, in each case, we are able to discover almost the same number of unmasked dimensions, independent of the starting point. It is also observed that the Wasserstein distance is minimized at the point where the mask reaches the optimal point (Refer to supplementary material for the plots).

%value of various losses (Section~\ref{sec:model}) as the training proceeds. 

%The Wasserstein distance becomes zero when $11$ out of $110$ dimensions are active on MNIST. Also the reconstruction error decreases initially and starts to increase when the mask values are not settled finally stabilize when the number of active dimensions become $11$. Also the FID score stabilizes at that point.
%It was observed that the FID score is least when the latent dimension matches the dimension of the data generation latent space. This experiment strengthens our claim further.

\begin{figure*}[ht]
        \begin{center}             
            \includegraphics[width=0.95\linewidth, keepaspectratio]{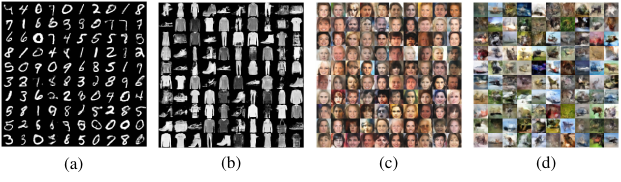}
        \end{center}
        \vspace{-4mm}
        \caption{Randomly generated (no cherry picking) images of (a) MNIST, (b) Fashion MNIST, (c) CelebA, and (d) CIFAR-10 datasets.}
        \vspace{-4mm}
        \label{fig:samples}
\end{figure*}

%Next, we empirically validate the effectiveness of our proposed algorithm. For this, we again take help of our synthetically created dataset and MNIST dataset. It was observed starting from $m=10$ to $m=110$, irrespective of the model capacity, the number of active dimensions at the completion of training is always in the range $[9, 14]$. For the MNIST dataset too, the FID score vs latent capacity curve is much flatter for our proposed model as seen in Figure \ref{fig:MAAE_VS_WAE}. For the synthetically created dataset,  the number of active dimensions is either exactly equal or differ by $1$ as compared to the true latent space dimension, $n$. 

% \begin{figure*}[!t]
%     \begin{center}             
%         \includegraphics[width=\linewidth, keepaspectratio]{./four_plots}
%     \end{center}
%     \caption{In MAAE experiment for latent capacity $110$ on MNIST dataset (a) Reconstruction Error, (b) Wasserstein Distance, and (c) FID Score are plotted against iteration.}
%     \label{fig:fourPlots}
% \end{figure*}

Finally, to measure generation quality, we present the FID scores of our method in Table~\ref{table:results} along with several  state-of-the-art AE-based models mentioned in section 2.
Our approach achieves the best FID score on all the datasets compared to the state-of-the-art AE based generative models. Performance of MAAE is also comparable to that of GANs listed in \cite{lucic2018gans}, despite using a simple norm based reconstruction loss and an isotropic uni-modal Gaussian prior. Figure~\ref{fig:samples} presents some randomly generated samples by our algorithm for each of the datasets. 

% table on co-varinace goes here.
The better FID scores of MAAE can be attributed to better distribution matching in the latent space between $\Psi(\vz)$ and $\Pi(\vz)$. But quantitatively comparing the matching between the two distributions is not easy as MAAE might mask out some of the latent dimensions resulting in a mismatch between the dimensionality of latent space in different models, thus rendering the usual metrics not suitable. We therefore calculate the averaged off-diagonal normalized absolute co-variance\footnote{Refer supplementary material for mathematical formula.} (NAC) of the encoded latent vectors and report it in Table \ref{table:covariancecomptable} (Refer supplementary material for the full Co-variance matrix). Since $\Pi(\vz)$ is assumed to be an isotropic Gaussian, ideally NAC should be zero and any deviation from zero indicates a mismatch. For MAAE we use only the unmasked latent dimensions for the NAC calculation, this is to ensure that NAC is not underestimated by considering the unused dimension. It is observed that for the same model capacity, MAAE has lesser NAC than the corresponding WAE indicating better distribution matching in the latent space.

\begin{table}[ht]
    \begin{center}
	\resizebox{\columnwidth}{!}{%
        \begin{tabular}{l r r r r}
            \hline
            & MNIST & Fashion & CIFAR-10 & CelebA \\
            \hline\hline
            \centered{VAE (cross-entr.)}    &   \centered{$16.6$}   &   \centered{$43.6$}   &   \centered{$106.0$}  &   \centered{$53.3$}   \\
            \centered{VAE (fixed variance)}   &   \centered{$52.0$} & \centered{$84.6$}   &   \centered{$160.5$}  &   \centered{$55.9$}    \\
            \centered{VAE (learned variance)}   &   \centered{$54.5$}    &   \centered{$60.0$}   &   \centered{$76.7$}   &   \centered{$60.5$}      	     \\
            \centered{VAE + Flow}   &   \centered{$54.8$}   &     \centered{$62.1$} &    \centered{$81.2$}  &     \centered{$65.7$}      	     \\
            \centered{WAE-MMD}  &   \centered{$115.0$}  &   \centered{$101.7$}  &    \centered{$80.9$}  &   \centered{$62.9$}      	     \\
            \centered{WAE-GAN}  &   \centered{$12.4$}  &   \centered{$31.5$}  &    \centered{$93.1$}  &   \centered{$66.5$}      	     \\
            \centered{2-Stage VAE}  &   \centered{$12.6$}   &   \centered{$29.3$}   &   \centered{$72.9$}   &   \centered{$44.4$}      	     \\
	        \centered{MAAE}    &   \centered{$\boldsymbol{10.5}$} & \centered{$\boldsymbol{28.4}$}	&	\centered{$\boldsymbol{71.9}$}	&	 \centered{$\boldsymbol{40.5}$}		     \\
            \hline
        \end{tabular}
	}
    \end{center}
    \vspace{-2mm}
    \caption{FID scores for generated images from different AE-based generative models (Lower is better).}
    \vspace{-2mm}
    \label{table:results}
\end{table}
\begin{table}[ht]
    \begin{center}
	\resizebox{\columnwidth}{!}{
        \begin{tabular}{l c c c c c }
            \hline
            Dataset & Model Capacity & \multicolumn{2}{c}{WAE} &  \multicolumn{2}{c}{MAAE}\\
            & &$m_{A}$ & NAC & $m_{A}$ & NAC\\
            \hline \hline
            
            $\text{Synthetic}_{8}$ & $16$ & $16$ & $0.040$ & $9$ & $\boldsymbol{0.030}$\\
            $\text{Synthetic}_{16}$ & $32$ & $32$ & $0.031$ & $16$ & $\boldsymbol{0.013}$\\
            MNIST & $64$ & $64$ & $0.027$ & $13$ & $\boldsymbol{0.020}$\\
            FMNIST & $128$ & $128$ & $0.025$ & $40$ & $\boldsymbol{0.019}$\\
            CIFAR-$10$ & $256$ & $256$ & $0.017$ & $120$ & $\boldsymbol{0.013}$\\
            CelebA & $256$ & $256$ & $0.046$ & $77$ & $\boldsymbol{0.039}$\\
            
            \hline
        \end{tabular}
        }
    \end{center}
    \vspace{-2mm}
    \caption{Average off-diagonal covariance NAC for both WAE and MAAE. $m_{A}$ represents the number of unmasked latent dimensions in the trained model. It is seen that MAAE has lower NAC values indicating lesser deviation of $\Psi(\vz)$ from $\Pi(\vz)$ as compared to a WAE.}
    \vspace{-2mm}
    \label{table:covariancecomptable}
\end{table}
These results clearly demonstrate that not only MAAE can achieve the best FID scores on a number of benchmarks datasets, it also serves as a first step in discovering the underlying latent structure for a given dataset. To the best of our knowledge, this is the first study analyzing (and discovering) the effect of latent dimensions on the generation quality.

%% file: discussion.tex
\section{DISCUSSION AND CONCLUSION}
Despite demonstrating its pragmatic success, we critically analyze the possible deviations of the practical cases from the presented analysis. 
More often than not, the naturally occurring data contains some noise superimposed onto the actual image. Thus, theoretically one can argue that this noise can be utilized to minimize the divergence between the distributions. Practically, however, this noise has a very low amplitude, so it can only work for a few extra dimensions, giving a slight overestimate of $n$. Further, in practice, not all latent dimensions contribute equally to the data generation. Since the objective of our model is to ignore noise dimensions, it may at times end up throwing away meaningful data dimensions which do not contribute significantly. This can lead to a slight underestimate of $n$ (which is occasionally observed during experimentation). Finally, neural networks, however deep, can represent only a certain level of complexity in a function which is simultaneous advantageous and otherwise. It is good because while we have shown that certain losses cannot be made zero for $m\neq n$, universal approximators can bring them arbitrarily close to zero, which is practically the same thing. Due to their limitation, however, we end up getting a U-curve. It is a disadvantageous because even at the $m\geq n$, the encoder and decoder networks might be unable to learn the appropriate functions, and at $m\leq n$, the Discriminator fails to make distributions apart. This implies that instead of discovering the exact same number of dimensions every time, we might get a range of values near the true latent dimension. Also, the severity of this problem is likely to increase with the complexity of the dataset (again corroborated by the experiments).\par 

To conclude, in this work, we have taken a step towards constructing an optimal latent space for improving the generation quality of Auto-Encoder based neural generative model. We have argued that, under the assumption two-step generative process, the optimal latent space for the AE-model is one where its dimensionality matches with that of the latent space of the generative process. Further, we have proposed a practical method to arrive at this optimal dimensionality from an arbitrary point by masking the `spurious' dimensions in AE-based generative models. Finally, we have shown the effectiveness of our method in improving the generation quality using several experiments on synthetic and real datasets.

%% file: supp_theory.tex
\section{THEORY}
% \subsection{Lemma 1 and Its Proof}
% % \begin{theorem}
% \noindent
% \textbf{Lemma 1}: A continuous function $\phi:\real{\alpha}\to\real{\beta}$ cannot have a continuous left inverse if $\alpha>\beta$.\\\\
% \textbf{Proof}: It follows trivially from the fact that such a function would define a homeomorphism from $\real{\alpha}$ to a subset of  $\real{\beta}$, whereas it is well known that these two spaces are not homeomorphic.\qed
% \end{theorem}

\subsection{DERIVATIONS FOR R1 AND R2}
In the main paper, we have stated the following conditions as requirements for optimal generation:
\begin{enumerate}
\item[R1] $ f(\widetilde\vz)=g'(g(f(\widetilde\vz)))\;\forall\  \widetilde\vz \ \in \real{n}$. 
    %\item[R2] $\not\exists h:\real{m}\to\{0,1\}$, such that $\int_{\real{m}}h(\vx})d\vx=0$ and $h(g(f(\vx)))=1\;\forall \vx\in \real{n}$.{\color{teal} this requirement is not clear? Is it ever used in the argument below?}{\color{red} basically that range of g(f(x)) has measure 0 in its output space. Also, this h is exactly what the discriminator needs to learn in order to win the adverserial game. It is used in the second last paragraph.} ({\color{teal} I see but we have to motivate this assumption in a different way because we are not using any reference to AAE here, this is a general section about AE and that we need matching of dimensions})
    \item[R2]$\mathcal{H}(\Psi,\Pi)$ on $\mathcal{Z}$ is minimal.
\end{enumerate}
In this section, we shall show that these are indeed necessary and sufficient to minimise the cross-entropy between the true data distribution $\Gamma$ and the generated data distribution $\Gamma'$.
\\Since auto-encoder based frameworks work through a latent space $\mathcal{Z}$, their objective is to minimise the cross entropy between the joint distribution of $\vx$ and $\vz$. We define these joint distributions as:
\begin{align}
    \begin{split}
        \Gamma(\vx,\vz)&=\Upsilon(\vx)\delta(g(\vx)=\vz)\\
        \Gamma'(\vx,\vz)&=\Pi(\vz)\delta(\vx=g'(\vz))\\
    \end{split}
\end{align}
where $\delta$ is the Dirac delta function. The cross entropy between these two distributions can be broken down as follows:
\begin{align}
    \begin{split}
        \mathcal{L}(\Gamma,\Gamma')&=\mathop{\mathbb{E}}_{(\vx,\vz)\sim\Gamma}(-\log(\Gamma'(\vx,\vz)))\\
        &=\mathop{\mathbb{E}}_{(\vx,\vz)\sim\Gamma}(-\log(\Gamma'(\vx|\vz)))\\
        &\quad+\mathop{\mathbb{E}}_{(\vx,\vz)\sim\Gamma}(\log(\frac{1}{\Pi(\vz)}))
    \end{split}
\end{align}
The first term in the final expression can further be expressed as:
\begin{align}
\begin{split}
    &\quad\mathop{\mathbb{E}}_{(\vx,\vz)\sim\Gamma}(-\log(\Gamma'(\vx|\vz)))\\&=-\int\limits_\vx\int\limits_\vz\Upsilon(\vx)\delta(g(\vx)=\vz)\log(\delta(\vx=g'(\vz)))d\vz d\vx\\
    &=-\int\limits_\vx\Upsilon(\vx)\log(\delta(\vx=g'(g(\vx))))d\vx
    \end{split}
\end{align}
If the equality inside the delta function does not hold at any point, it will push the logarithm to negative infinity, and in turn the entire quantity will become very high. To prevent this, we need $\vx=g'(g(\vx))$ at all points. Since $\vx$ varies over the range of $f$, this reduces to R1.\\
In the second term, the expectation is over a joint distribution, but the variable $\vx$ never appears inside the expectation, so it is safe to take the expectation over the marginal of $\vz$. However, this marginal, $\Gamma(\vz)$ is exactly the distribution imposed on the latent space by the encoder and the data distribution, which we previously called $\Psi$. Making this change turns this term into $\mathcal{H}(\Psi,\Pi)$ and since we need this to be minimal, we recover the R2.\\

\subsection{DISCUSSION}

Our generative process assumes that $n$, the dimension of the input space of $f$ is minimal. Intuitively this means that each of the latent dimension contributes in generation of some (possibly small)
region in the domain of the observed data but it is not necessary that every latent dimension (independently) affects every observed data point.

%Intuitively this means that each of the latent dimension contributes in generation of some small (albeit infinitesimal) region in the domain of the observed data but not that every latent dimension affect every observed data point.

For example, consider the case where a leaf is being photographed. A young leaf in broad daylight has colour roughly (120,100,50) in the HSL system. As the age of the leaf increases, the lightness starts to fall, but a similar fall in lightness will also be observed with fading daylight. At this point, lighting conditions and age of leaf have identical effect on the appearance of the leaf. However, after a point, age will start reducing the hue of the leaf, while lighting conditions will continue decreasing its lightness. Since at this point these two factors influence the outcome differently, they can be separate factors in our input space. On the other hand, the distance from which the photo was taken and the optical zoom of the lens will always have similar effect, and therefore, only one of these is allowed as a factor. Note that this example is presented for illustrative purposes only, and in real cases, the input factors are unlikely to directly map to real-world causes.

\subsection{INTUITION FOR LEMMA 1}
In the main paper, we have claimed that given a set $S$, defined as:
$S$={\large\{}$(\frac{a_0+0.5}{\epsilon},\ldots,\frac{a_{n-1}+0.5}{\epsilon})\big|a_i\in\{0,\ldots,\epsilon-1\}${\large \}}
if we build closed balls around each point in $S$, then every point in $[0,1]^n$ lies in atleast one of these balls. To further the intuition behind this, we present an illustration for the case where $n=2$ and $\epsilon=4$.\\
\begin{figure}[ht]\label{intuit}
    \begin{center}
         \includegraphics[width=\linewidth, keepaspectratio]{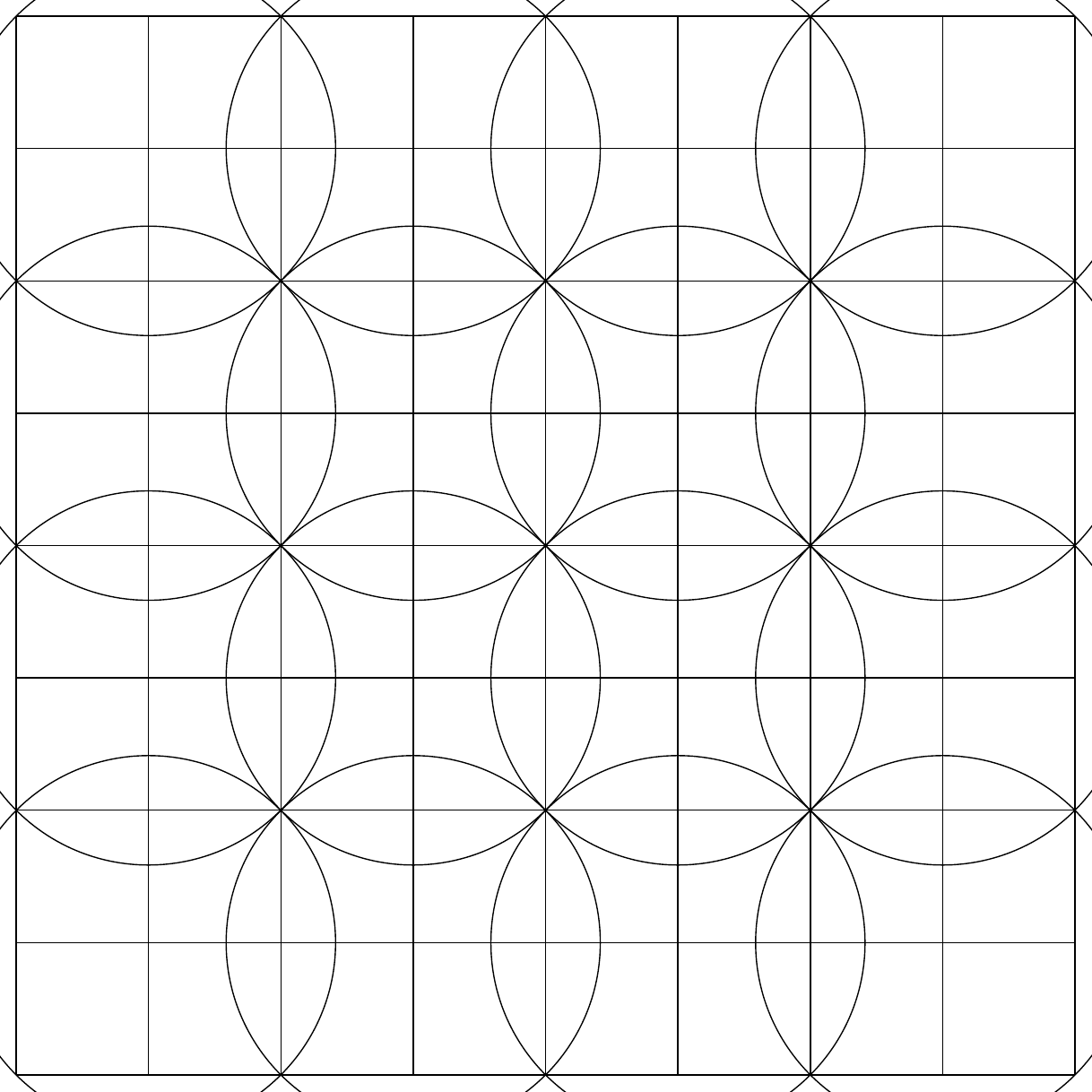}
    \end{center}
    \vspace{-2mm}
    \caption{For $n=2$, $\epsilon = 4$ radius of each ball, $r=\frac{\sqrt(2)}{8}$.}
    \vspace{-4mm}
    \label{fig:intuition}
\end{figure}
\par In figure \ref{fig:intuition} the big square represents the unit square. By the nature of Cartesian space, this can be tiled completely by 64 smaller squares having side $\frac{1}{8}$ units. The 16 circles represent the closed balls in $\mathbb{R}^2$. It is easy to see that each of the smaller squares lies completely in some circle. Since each point in the bigger square must lie in one of the smaller squares, they also lie in one of the circles.

%% file: supp_model.tex
\section{OBJECTIVE, TRAINING AND ARCHITECTURE OF MAAE}\label{sec:supp_model}
In this section we describe the MAAE model and the training algorithm in detail.

\begin{enumerate}
    \item {\bf Re-construction Pipeline}: This is the standard pipeline in any given AE based model, which tries to minimize the reconstruction loss (R1). An input sample $\vx$ is passed through the encoder \encoder results in $\hat{\boldsymbol{z}}$, the corresponding representation in the latent space. The new addition here is the Hadamard product with the mask $\mu$ (explained next), resulting in the masked latent space representation $\mu \odot \hat{\vz}$. The masked representation is then fed to the decoder \decoder to obtain the re-constructed output $\hat{\vx}$. The goal here is to minimize the norm of the difference between $\vx$ and $\hat{\vx}$.
    \item {\bf Masking Pipeline}: Introduction of a mask is one of the novel contributions of our work, and this is the second part of our architecture presented in the middle of the Figure~\ref{fig:architecture}. Our mask is represented as $\mu$ and is a binary vector of size $m$ (model capacity). Ideally, the mask would be a binary vector, but in order to make it learnable, we relax it to be continuous valued, while imposing certain regularizers so that it does not deviate too much from 0 or 1 during learning.  
    \item {\bf Distribution-Matching Pipeline:} This is the third part of our architecture presented at the bottom of Figure~\ref{fig:architecture}. Objective of this pipeline is to minimize the distribution loss between a prior distribution, $\Pi$, and the distribution $\Psi$ imposed on the latent space by the encoder. $\vz$ is a random vector sampled from the prior distribution, whose Hadamard product is taken with the mask $\mu$ (similar to in the case of encoder), resulting in a masked vector $\mu \odot \vz$. This masked vector is then passed through the network \discriminator, where the goal is to separate out the samples coming from prior distribution ($\vz$) from those coming from the encoded space ($\hat{\vz}$) using some divergence metric. We use the principles detailed in \cite{arjovsky2017wasserstein} using the Wasserstein's distance to measure the distributional divergence. Note that $H_{\zeta}$ has two inputs namely, samples of $\Pi(\vz)$ and output of $E_{\kappa}$.
\end{enumerate}

\subsection{OBJECTIVE FUNCTIONS OF MAAE}
Next, corresponding to each of the components above, we present a loss function where $s$ represents the batch size.
\begin{enumerate}
    \item {\bf Auto-Encoder Loss:} This is the standard loss to capture the quality of re-construction as used earlier in the AE literature. In addition, we have a term corresponding to minimization of the variance over the masked dimensions in the encoded output in a batch. The intuition is that encoder should not inject information into the dimensions which are going to be masked anyway. The loss is specified as:  
        \begin{equation}\begin{split}
            L_{ae} &= \frac{\alpha_1}{s}\sum_{i=1}^{s}||\vx^{(i)}-D_\psi(\mu \odot E_\kappa(\vx^{(i)}))|| \\
            &\qquad+ \alpha_2(\delta^TDiag(A))
            %CoVar(E_\kappa(X))))
        \end{split}\end{equation}
    $A$ represents the co-variance matrix for the encoding matrix $E_{\kappa}(X)$, $X$ being the data matrix for the current batch. $\delta$ is the vector obtained by applying the function $a(u)=e^{-\gamma \times u}$ point-wise to $\mu$. $\alpha_1$, $\alpha_2$ and $\gamma$ are hyperparameters.
    %is a hyperparameter.
    %where $a(\mu)=e^{-\gamma \times \mu_{i}^{2}}\;\; \forall \;\;i$. %\in \{1,2,\ldots,z\}$
    
    \item {\bf Generator Loss:} This is the loss capturing the quality of generation in terms of how far the generated distribution is from the prior distribution. This loss measures the ability of the encoder to generate the samples such that they are coming from $\Pi(\rz)$ which is ensured using the generator loss mentioned in \cite{arjovsky2017wasserstein}:
    \begin{equation}
            L_{gen} = -\frac{1}{s}\sum_{i=1}^{s}H_\zeta(\mu \odot E_\kappa (\vx^{(i)}))
    \end{equation}
        
    \item {\bf Distribution-Matching Loss:} This is the loss incurred by the Distribution-matching network, \discriminator in matching the distributions. We use Wasserstein's distance (\cite{arjovsky2017wasserstein}) to measure the distributional closeness with the following loss: 
    \begin{equation}\begin{split}
            L_{dm} &= -\frac{1}{s}\sum_{i=1}^{s}H_\zeta( \mu \odot \vz^{(i)}) %\\
                      %&\qquad 
                      +\frac{1}{s}\sum_{i=1}^{s}H_\zeta (\mu \odot \hat{\vz}^{(i)}) \\
                      %E_\kappa (\vx^{(i)})) \\
                      &\qquad + 
                  \frac{\beta_2}{s}\sum_{i=1}^{s}\big(\lvert\lvert\nabla_{\vz_{avg}}^{(i)}H_\zeta (\mu \odot \vz_{avg}^{(i)})\lvert\lvert - 1\big)^2
        \end{split}\end{equation}
 Recall that $\hat{\vz}^{(i)} = E_{\kappa}(\vx^{(i)})$. Further, we have used $\vz_{avg}^{(i)} = \beta_1\vz^{(i)} + (1-\beta_1)\hat{\vz}^{(i)}$. $\beta_1,\beta_2$ are hyper parameters, with $\beta_1 \sim \mathcal{U}[0, 1]$, and $\beta_2$ set as in  (\cite{gulrajani2017improved}).
    \item {\bf Masking Loss:} This is the loss capturing the quality of the current mask. The loss is a function of three terms (1) Auto-encoder loss (2) distribution matching loss (3) a regularizer to ensure that $\mu$ parameters stay close to $0$ or $1$. This can be specified as:
        \begin{equation}\begin{split}
            L_{mask} &= \frac{\lambda_1}{s}\sum_{i=1}^{s}||\vx^{(i)}-D_\psi(\mu \odot E_\kappa(\vx^{(i)}))|| \\
            &\qquad+ \lambda_2(1 + \omega)^2 
            %&\qquad
            + \lambda_3 \sum_{j=1}^{m}|\mu_j(\mu_j - 1)|
        \end{split}\end{equation}
        where 
        %\begin{equation}\begin{split}
            $\omega = \frac{1}{s}\sum_{i}H_\zeta(\mu \odot \vz^{(i)})
                 - \frac{1}{s}\sum_{i}H_\zeta(\mu \odot E_\kappa (\vx^{(i)}))$ is the Wasserstein's distance. 
        %\end{split}\end{equation}
Here $\lambda_1$ and $\lambda_2$ and $\lambda_3$ are hyper-parameters (Supp. material). 
% $\odot$ represents Hadamard product between two vectors. 
\end{enumerate}

\subsection{TRAINING ALGORITHM}
During training, we optimize each of the four losses specified above in turn. Specifically, in each learning loop, we optimize the $L_{ae}$, $L_{dm}$, $L_{gen}$ and $L_{mask}$, in that order using a learning schedule. 
We use RMSProp for our optimization as described in algorithm \ref{alg:maae-train-loop}.

 \begin{algorithm}
    \caption{Pseudo code for the training loop of MAAE \label{alg:maae-train-loop}}
    \hspace*{\algorithmicindent} \textbf{Hyper-parameters:} $\alpha_1 = 1$, $\alpha_2 = 100$, $\gamma = 10$, $\lambda_1 = 1000$, $\lambda_2 = 1$ and $\lambda_3 = \frac{2}{m}$ \\
    \begin{algorithmic}[1]
        \Function{Train}{}
            \Let{$\lambda_3$}{$\frac{2}{m}$}
            \For{$i \gets 1 \textrm{ to } training\_steps$}
                \For{$j \gets 1 \textrm{ to } ae\_training\_ratio$}
                    \Let{$t$}{$ij+j$}
                    \Let{$g_{\kappa}(t)$}{$\nabla_\kappa L_{ae}$}
                    \Let{$\kappa_{t}$}{$\kappa_{t-1}-\frac{\eta_{ae} g_{\kappa}(t)}{\sqrt{\rho g_{\kappa}(t) + (1-\rho)g_{\kappa}^2(t-1)+\epsilon}}$}
                    \Let{$g_{\Psi}(t)$}{$\nabla_\Psi L_{ae}$}
                    \Let{$\Psi_{t}$}{$\Psi_{t-1}-\frac{\eta_{ae} g_{\Psi}(t)}{\sqrt{\rho g_{\Psi}(t) + (1-\rho)g_{\Psi}^2(t-1)+\epsilon}}$}
                    
                \EndFor
                \For{$j \gets 1 \textrm{ to } disc\_training\_ratio$}
                    \Let{$t$}{$ij+j$}
                    \Let{$g_{\zeta}(t)$}{$\nabla_\zeta L_{disc}$}
                    \Let{$\zeta_{t}$}{$\zeta_{t-1}-\frac{\eta_{disc} g_{\zeta}(t)}{\sqrt{\rho g_{\zeta}(t) + (1-\rho)g_{\zeta}^2(t-1)+\epsilon}}$}
                \EndFor
                \Let{$g_{\kappa}(i)$}{$\nabla_\kappa L_{gen}$}
                \Let{$\kappa_{i}$}{$\kappa_{i-1}-\frac{\eta_{gen} g_{\kappa}(i)}{\sqrt{\rho g_{\kappa}(i) + (1-\rho)g_{\kappa}^2(i-1)+\epsilon}}$}
                \If {$i \% reg\_schedule\_interval == 0$}
                    \Let{$\lambda_3$}{$\lambda_3 \times 2$}
                \EndIf
                \Let{$g_{M}(i)$}{$\nabla_M L_{mask}$}
                \Let{$M_{i}$}{$M_{i-1}-\frac{\eta_{mask} g_{M}(i)}{\sqrt{\rho g_{M}(i) + (1-\rho)g_{M}^2(i-1)+\epsilon}}$}
            \EndFor
        \EndFunction
    \end{algorithmic}
\end{algorithm}

\subsection{ARCHITECTURE FOR SYNTHETIC dataset}
Here we provide the detailed architecture of \encoder \decoder and \discriminator for different experiments performed in this work.
\subsubsection{$E_{\kappa}$}
\begin{equation*}
    \begin{split}
        & \boldsymbol{x}\in\mathbb{R}^{128} \\
        &\to \text{FC}_{1000} \to \text{ReLU} \\
        &\to \text{FC}_{1000} \to \text{ReLU} \\
        &\to \text{FC}_{1000} \to \text{ReLU}  \\
        &\to \text{FC}_{1000} \to \text{ReLU}  \\
        &\to \text{FC}_{1000} \to \text{ReLU}  \\
        &\to \text{FC}_{m}
    \end{split}
\end{equation*}

\subsubsection{$D_{\psi}$}
\begin{equation*}
    \begin{split}
        & \boldsymbol{z}\in\mathbb{R}^{m} \\
        &\to \text{FC}_{1000} \to \text{ReLU} \\
        &\to \text{FC}_{1000} \to \text{ReLU} \\
        &\to \text{FC}_{1000} \to \text{ReLU}  \\
        &\to \text{FC}_{1000} \to \text{ReLU}  \\
        &\to \text{FC}_{1000} \to \text{ReLU}  \\
        &\to \text{FC}_{m}
    \end{split}
\end{equation*}

\subsubsection{$H_{\zeta}$}
\begin{equation*}
    \begin{split}
        & \boldsymbol{z}\in\mathbb{R}^{m} \\
        &\to \text{FC}_{1000} \to \text{ReLU} \\
        &\to \text{FC}_{1000} \to \text{ReLU} \\
        &\to \text{FC}_{1000} \to \text{ReLU}  \\
        &\to \text{FC}_{1000} \to \text{ReLU}  \\
        &\to \text{FC}_{1000} \to \text{ReLU}  \\
        &\to \text{FC}_{1}
    \end{split}
\end{equation*}

\subsection{ARCHITECTURE FOR REAL dataset}
\subsubsection{MNIST}
\subsubsubsection{$E_{\kappa}$}
\begin{equation*}
    \begin{split}
        & \boldsymbol{x}\in\mathbb{R}^{28\times28\times1} \\
        &\to \text{FC}_{1024} \to \text{ReLU} \\
        &\to \text{FC}_{1024} \to \text{ReLU} \\
        &\to \text{FC}_{1024} \to \text{ReLU}  \\
        &\to \text{FC}_{1024} \to \text{ReLU}  \\
        &\to \text{FC}_{m}
    \end{split}
\end{equation*}
\subsubsubsection{$D_{\psi}$}
\begin{equation*}
    \begin{split}
        & \boldsymbol{z}\in\mathbb{R}^{m} \\
        &\to \text{FC}_{1024} \to \text{ReLU} \\
        &\to \text{FC}_{1024} \to \text{ReLU} \\
        &\to \text{FC}_{1024} \to \text{ReLU} \\
        &\to \text{FC}_{1024} \to \text{ReLU} \\
        &\to \text{FC}_{28 \times 28} \to \text{Sigmoid} \\
        &\to \text{Reshape}_{28 \times 28}
    \end{split}
\end{equation*}
\subsubsubsection{$H_{\zeta}$} \label{mnistDisc}
\begin{equation*}
    \begin{split}
        & \boldsymbol{z}\in\mathbb{R}^{m} \\
        &\to \text{FC}_{1024} \to \text{ReLU} \\
        &\to \text{FC}_{1024} \to \text{ReLU} \\
        &\to \text{FC}_{1024} \to \text{ReLU} \\
        &\to \text{FC}_{1024} \to \text{ReLU} \\
        &\to \text{FC}_{1}
    \end{split}
\end{equation*}

\subsubsection{Fashion MNIST}
\subsubsubsection{$E_{\kappa}$}
\begin{equation*}
    \begin{split}
        & \boldsymbol{x}\in\mathbb{R}^{28\times28\times1} \\
        &\to \text{CONV}_{64; k=(4, 4); s=(2, 2)} \to \text{BN} \to \text{ReLU} \\
        &\to \text{CONV}_{128; k=(4, 4); s=(2, 2)} \to \text{BN} \to \text{ReLU} \\
        &\to \text{FC}_{1024} \to \text{BN} \to \text{ReLU} \\
        &\to \text{FC}_{m}
    \end{split}
\end{equation*}
\subsubsubsection{$D_{\psi}$}
\begin{equation*}
    \begin{split}
        & \boldsymbol{z}\in\mathbb{R}^{m} \\
        &\to \text{FC}_{1024} \to \text{BN} \to \text{ReLU} \\
        &\to \text{FC}_{7 \times 7 \times 128} \to \text{BN} \to \text{ReLU} \\
        &\to \text{Reshape}_{7 \times 7 \times 128} \\
        &\to \text{TCONV}_{128; k=(4, 4); s=(2, 2)} \to \text{BN} \to \text{ReLU} \\
        &\to \text{TCONV}_{128; k=(4, 4); s=(2, 2)} \to \text{BN} \to \text{ReLU} \\
        &\to \text{CONV}_{1; k=(3, 3); s=(1, 1)} \to \text{Sigmoid}
    \end{split}
\end{equation*}
\subsubsubsection{$H_{\zeta}$}
Same as in \ref{mnistDisc}

\subsubsection{CIFAR-$10$}
\subsubsubsection{$E_{\kappa}$}
\begin{equation*}
    \begin{split}
        & \boldsymbol{x}\in\mathbb{R}^{32\times32\times3} \\
        &\to \text{CONV}_{128; k=(4, 4); s=(2, 2)} \to \text{BN} \to \text{ReLU} \\
        &\to \text{CONV}_{128; k=(4, 4); s=(2, 2)} \to \text{BN} \to \text{ReLU} \\
        &\to \text{CONV}_{256; k=(4, 4); s=(2, 2)} \to \text{BN} \to \text{ReLU} \\
        &\to \text{FC}_{1024} \to \text{BN} \to \text{ReLU} \\
        &\to \text{FC}_{1024} \to \text{BN} \to \text{ReLU} \\
        &\to \text{FC}_{1024} \to \text{BN} \to \text{ReLU} \\
        &\to \text{FC}_{1024} \to \text{BN} \to \text{ReLU} \\
        &\to \text{FC}_{m}
    \end{split}
\end{equation*}
\subsubsubsection{$D_{\psi}$}
\begin{equation*}
    \begin{split}
        & \boldsymbol{z}\in\mathbb{R}^{m} \\
        &\to \text{FC}_{2\times2\times512} \to \text{BN} \to \text{ReLU} \\
        &\to \text{Reshape}_{2 \times 2 \times 512} \\
        &\to \text{TCONV}_{256; k=(4, 4); s=(2, 2)} \to \text{BN} \to \text{ReLU} \\
        &\to \text{CONV\_RES\_BLOCK}_{256} \to \text{BN} \to \text{ReLU} \\
        &\to \text{CONV\_TCONV}_{256; k=(4, 4); s=(2, 2)} \to \text{BN} \to \text{ReLU} \\
        &\to \text{CONV\_RES\_BLOCK}_{256} \to \text{BN} \to \text{ReLU} \\
        &\to \text{TCONV}_{256; k=(4, 4); s=(2, 2)} \to \text{BN} \to \text{ReLU} \\
        &\to \text{CONV\_RES\_BLOCK}_{256} \to \text{BN} \to \text{ReLU} \\
        &\to \text{TCONV}_{256; k=(4, 4); s=(2, 2)} \to \text{BN} \to \text{ReLU} \\
        &\to \text{CONV\_RES\_BLOCK}_{256} \to \text{BN} \to \text{ReLU} \\
        &\to \text{CONV}_{3; k=(3, 3); s=(1, 1)} \to \text{Sigmoid}
    \end{split}
\end{equation*}
\subsubsubsection{$H_{\zeta}$}
Same as in \ref{mnistDisc}

\subsubsection{CelebA}
\subsubsubsection{$E_{\kappa}$}
\begin{equation*}
    \begin{split}
        & \boldsymbol{x}\in\mathbb{R}^{64\times64\times3} \\
        &\to \text{CONV}_{16; k=(3, 3); s=(1, 1)} \to \text{BN} \\
        &\to \text{CONV\_RES\_BLOCK}_{16} \to \text{CONV}_{32; k=(4, 4); s=(2, 2)} \\
        &\to \text{CONV\_RES\_BLOCK}_{32} \to \text{CONV}_{64; k=(4, 4); s=(2, 2)} \\
        &\to \text{CONV\_RES\_BLOCK}_{64} \to \text{CONV}_{64; k=(4, 4); s=(2, 2)} \\
        &\to \text{CONV\_RES\_BLOCK}_{64} \\
        &\to \text{FC\_RES\_BLOCK}_{512} \\
        &\to \text{FC}_{m}
    \end{split}
\end{equation*}

\begin{figure*}[ht]
    \begin{center}
         \includegraphics[width=\textwidth, keepaspectratio]{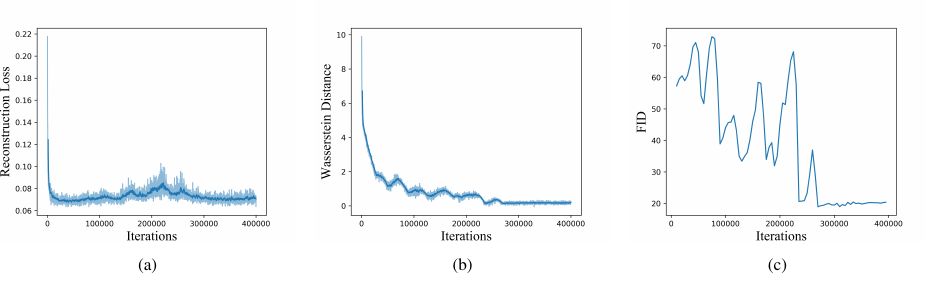}
    \end{center}
    \vspace{-2mm}
    \caption{(a) Reconstruction loss, (b) Wasserstein distance, and (c) FID plot w.r.t. training iterations for MAAE model on MNIST.}
    \vspace{-4mm}
    \label{fig:training_report}
\end{figure*}

\subsubsubsection{$D_{\psi}$}
\begin{equation*}
    \begin{split}
        & \boldsymbol{z}\in\mathbb{R}^{m} \\
        &\to \text{FC}_{2\times2\times16} \to \text{BN} \to \text{ReLU} \\
        &\to \text{Reshape}_{2 \times 2 \times 16} \\
        &\to \text{TCONV}_{32; k=(4, 4); s=(2, 2)} \to \text{BN} \to \text{ReLU} \\
        &\to \text{CONV\_RES\_BLOCK}_{32} \to \text{BN} \to \text{ReLU} \\
        &\to \text{TCONV}_{64; k=(4, 4); s=(2, 2)} \to \text{BN} \to \text{ReLU} \\
        &\to \text{CONV\_RES\_BLOCK}_{64} \to \text{BN} \to \text{ReLU} \\
        &\to \text{TCONV}_{128; k=(4, 4); s=(2, 2)} \to \text{BN} \to \text{ReLU} \\
        &\to \text{CONV\_RES\_BLOCK}_{128} \to \text{BN} \to \text{ReLU} \\
        &\to \text{TCONV}_{256; k=(4, 4); s=(2, 2)} \to \text{BN} \to \text{ReLU} \\
        &\to \text{CONV\_RES\_BLOCK}_{256} \to \text{BN} \to \text{ReLU} \\
        &\to \text{TCONV}_{512; k=(4, 4); s=(2, 2)} \to \text{BN} \to \text{ReLU} \\
        &\to \text{CONV\_RES\_BLOCK}_{512} \to \text{BN} \to \text{ReLU} \\
        &\to \text{CONV}_{3; k=(3, 3); s=(1, 1)} \to \text{Sigmoid}
    \end{split}
\end{equation*}
\subsubsubsection{$H_{\zeta}$}
Same as in \ref{mnistDisc}

%% file: supp_expt.tex
\section{EXPERIMENTAL RESULTS}
\subsection{DETAILS OF THE SYNTHETIC DATASET}
\begin{figure}[ht]
    \begin{center}
         \includegraphics[width=\linewidth, keepaspectratio]{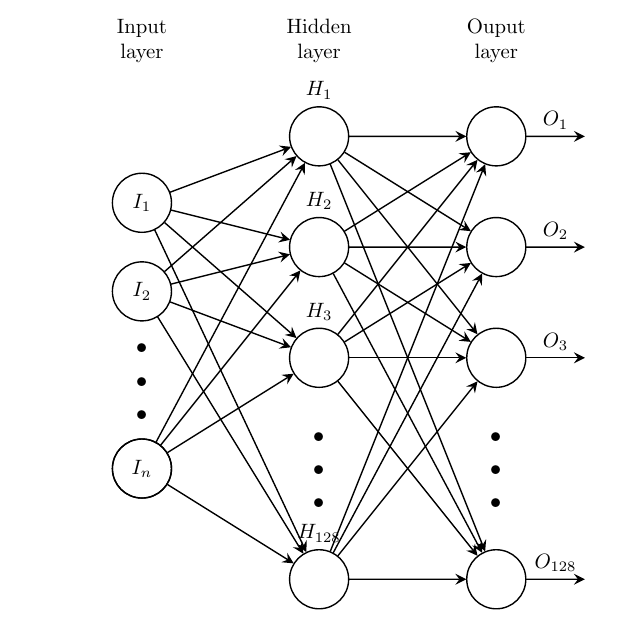}
    \end{center}
    \vspace{-2mm}
    \caption{Architecture for Synthetic Dataset generation.}
    \vspace{-4mm}
    \label{fig:toyGenArch}
\end{figure}

Figure \ref{fig:toyGenArch}, shows the architecture for synthetic data generation. The input layer has $n$ nodes, representing $n$ dimensions of the samples from a Gaussian distribution. The hidden layer and output layer introduce two layers of non-linearity and blow up the dimension from $n$ to $128$ of the synthetic dataset. In our experiments we have chosen $n=8$, and $n=16$.

\subsection{\uppercase{Analysis of Training of MAAE on MNIST}}
As discussed in section $5.2$ of the main paper, we can see in figure \ref{fig:training_report} (b) that for $m=110$, the Wasserstein distance becomes zero when number of active dimensions, $m_A = 11$ (Refer to Figure 5(c) in main paper). Also in Figure  \ref{fig:training_report} (a), and (c) we see the reconstruction error stabilizes at that point and the FID score becomes minimum.

\subsection{\uppercase{Normalised Absolute Co-variance Matrix: WAE vs MAAE}}
\begin{figure*}[ht]
    \begin{center}
         \includegraphics[width=\textwidth, keepaspectratio]{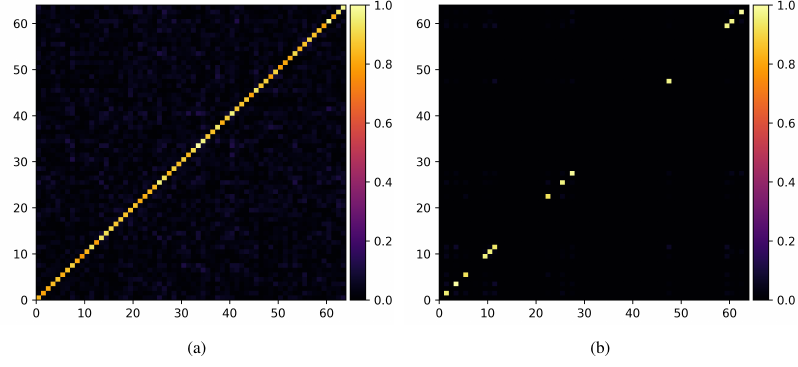}
    \end{center}
    \vspace{-2mm}
    \caption{Co-variance Matrix of (a) WAE (b) MAAE latent representation for MNIST dataset.}
    \vspace{-4mm}
    \label{fig:mnist_cov_mat}
\end{figure*}

\begin{figure*}[ht]
    \begin{center}
         \includegraphics[width=\textwidth, keepaspectratio]{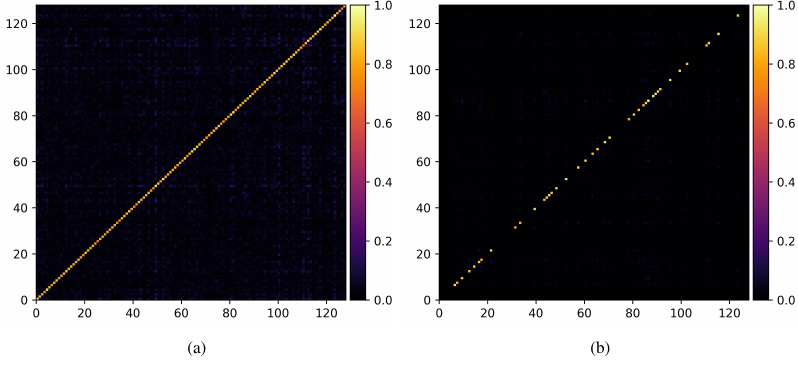}
    \end{center}
    \vspace{-2mm}
    \caption{Co-variance Matrix of (a) WAE (b) MAAE latent representation for Fashion MNIST dataset.}
    \vspace{-4mm}
    \label{fig:fashion_cov_mat}
\end{figure*}

\begin{figure*}[ht]
    \begin{center}
         \includegraphics[width=\textwidth, keepaspectratio]{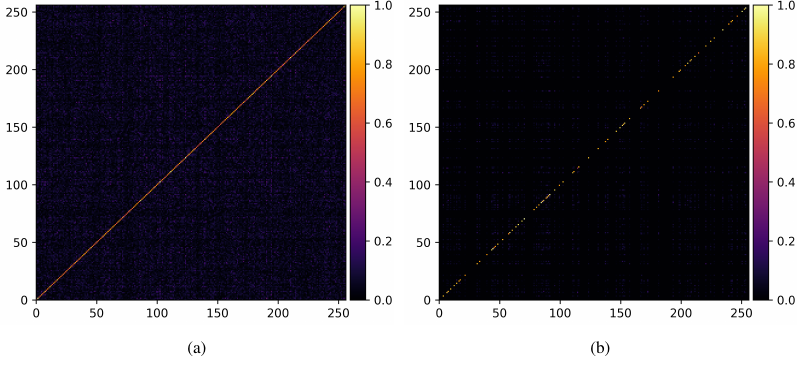}
    \end{center}
    \vspace{-2mm}
    \caption{Co-variance Matrix of (a) WAE (b) MAAE latent representation for CelebA dataset.}
    \vspace{-4mm}
    \label{fig:celeba_cov_mat}
\end{figure*}

\begin{figure*}[ht]
    \begin{center}
         \includegraphics[width=\textwidth, keepaspectratio]{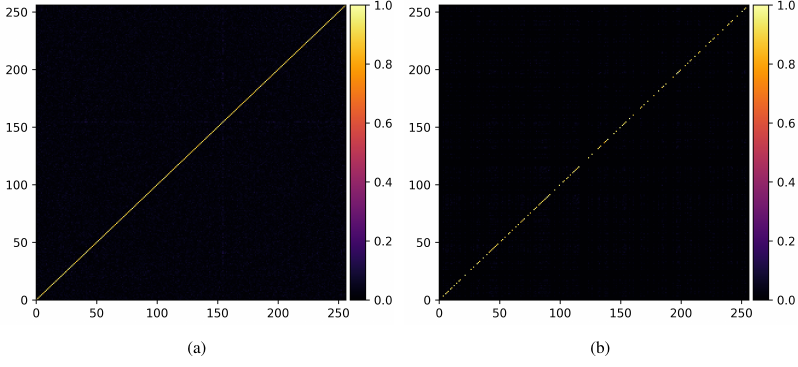}
    \end{center}
    \vspace{-2mm}
    \caption{Co-variance Matrix of (a) WAE (b) MAAE latent representation for CIFAR-$10$ dataset.}
    \vspace{-4mm}
    \label{fig:cifar10_cov_mat}
\end{figure*}

The following formula is used to compute the co-variance matrix over a batch size $b_s = 5000$.
\begin{equation}
\Sigma = \lvert\sum_{i=1}^{i=b_s}(\hat{\boldsymbol{z}}^{(i)} - \mu_{\hat{\boldsymbol{z}}})(\hat{\boldsymbol{z}}^{(i)} - \mu_{\hat{\boldsymbol{z}}})^T\lvert
\end{equation}
In figure \ref{fig:mnist_cov_mat}, \ref{fig:fashion_cov_mat}, \ref{fig:celeba_cov_mat}, and \ref{fig:cifar10_cov_mat} we have plotted the normalized co-variance matrix $\frac{\Sigma - min(\Sigma)}{max(\Sigma)-min(\Sigma)}$.
\par
Also from figure \ref{fig:mnist_cov_mat}, \ref{fig:fashion_cov_mat}, \ref{fig:celeba_cov_mat}, and \ref{fig:cifar10_cov_mat}, we see that the off-diagonal entries in the co-variance matrix corresponding to a masked dimension in MAAE model are very close to zero. Therefore, for a fair comparison of the average off-diagonal value in table $(2)$ of the main paper, we neglect those dimensions in the MAAE matrix.

